\pdfoutput=1

\documentclass[11pt, table]{article}
\usepackage{authblk}
\usepackage[]{acl}

\usepackage{times}
\usepackage{latexsym}
\usepackage[title]{appendix}
\usepackage{enumitem}
\usepackage{booktabs}
\usepackage{caption}
\usepackage{graphicx}
\usepackage{amsmath}
\usepackage{soul}
\usepackage{cleveref}
\usepackage{xcolor}
\definecolor{myorange}{RGB}{223, 130, 68}
\definecolor{myred}{RGB}{176, 35, 24}
\newcommand{\ctext}[3][RGB]{%
  \begingroup
  \definecolor{hlcolor}{#1}{#2}\sethlcolor{hlcolor}%
  \hl{#3}%
  \endgroup
}

\newcommand{\mypink}[1]{%
  \ctext[RGB]{247, 206, 205}{#1}
}
\newcommand{\myblue}[1]{%
  \ctext[RGB]{165,199,232}{#1}
}

\usepackage{amsfonts}
\usepackage{textcomp, xspace}
\usepackage{pbox}
\usepackage{array}
\newcolumntype{R}[1]{>{\raggedleft\let\newline\\\arraybackslash\hspace{0pt}}m{#1}}
\newcolumntype{L}[1]{>{\raggedright\let\newline\\\arraybackslash\hspace{0pt}}m{#1}}
\usepackage{tabu}
\usepackage{physics}
\usepackage{arydshln} 
\usepackage{multirow}
\usepackage{xcolor}
\usepackage[normalem]{ulem}
\useunder{\uline}{\ul}{}

\crefformat{section}{\S#2#1#3} 
\crefformat{subsection}{\S#2#1#3}
\crefformat{subsubsection}{\S#2#1#3}

\usepackage[T1]{fontenc}

\usepackage[utf8]{inputenc}

\usepackage{microtype}

\newcommand{\nop}[1]{}
\newcommand{\add}[1]{\textcolor{red}{#1}}

\usepackage{multirow}
\usepackage{amsmath}
\usepackage{makecell}
\usepackage{array}
\usepackage{diagbox}

\newcommand{\model}{\textsc{POLITICS}\xspace}

\newcommand{\data}{\textsc{BigNews}\xspace}

\newcommand{\datad}{\textsc{BigNewsBLN}\xspace}

\newcommand{\aligndata}{\textsc{BigNewsAlign}\xspace}

\let\svthefootnote\thefootnote
\newcommand\freefootnote[1]{%
  \let\thefootnote\relax%
  \footnotetext{#1}%
  \let\thefootnote\svthefootnote%
}

\title{\model: Pretraining with Same-story Article Comparison \\for Ideology Prediction and Stance Detection}

\author[1,\thanks{\quad Equal contribution by the first two authors.}]{\textbf{Yujian Liu}}
\author[1,*]{\textbf{Xinliang Frederick Zhang}}
\author[1]{\textbf{David Wegsman}}
\author[2]{\\\textbf{Nick Beauchamp}}
\author[1]{\textbf{Lu Wang}}

\affil[1]{Computer Science and Engineering, University of Michigan, Ann Arbor, MI}
\affil[2]{Department of Political Science, Northeastern University, Boston, MA}

\affil[1]{\{\texttt{yujianl,xlfzhang,dwegs,wangluxy\}@umich.edu}}
\affil[2]{\texttt{n.beauchamp@northeastern.edu}}

\begin{document}
\maketitle

\begin{abstract}
Ideology is at the core of political science research. Yet, there still does not exist general-purpose tools to characterize and predict ideology across different genres of text. 
To this end, we study Pretrained Language Models using novel ideology-driven pretraining objectives that rely on the comparison of articles on the same story written by media of different ideologies.
We further collect a large-scale dataset, consisting of more than $3.6$M political news articles, for pretraining.
Our model \model  outperforms strong baselines and the previous state-of-the-art models on ideology prediction and stance detection tasks. Further analyses show that \model is especially good at understanding long or formally written texts, and is also robust in few-shot learning scenarios. 

\end{abstract}
\section{Introduction}

Ideology is an ubiquitous factor in political science, journalism, and media studies \citep{mullins_1972, Michael_Freeden, martin2015ideology}.
Decades of work has gone into measuring ideology based on voting data~\citep{poole1985spatial,voteview}, survey results~\citep{preotiuc-pietro-etal-2017-beyond,ansolabehere2008strength,kim1998voter,gabel2000putting}, social networks~\citep{Barbera_jost}, campaign donation records \citep{Bonica2013MappingTI}, and textual data \citep{laver_benoit_garry_2003,diermeier2012language,Gentzkow_Shapiro,Volkens:2021}. Each of those approaches has its strengths and weaknesses. For instance, many political figures do not have voting records; surveys are expensive and politicians are often unwilling to disclose ideology.
By contrast, political text is abundant, ubiquitous, yet challenging to work with since language is complex in nature, often domain-specific, and generally unlabeled.
There thus remains a strong need for general-purpose tools for measuring ideology using text that can be applied across multiple genres. 

\begin{figure}
\small
\begin{tabular}{ p{72mm} }
\toprule
\textbf{News Story:} \textit{Donald Trump tests positive for COVID-19.}\\
\midrule
\textbf{Daily Kos} (left): It's now clear that Donald Trump \myblue{\textbf{lied}} to the nation about when he received a positive test for COVID-19.
$\ldots$
they’re continuing to act as if nothing has changed—and that \myblue{\textbf{disregarding science}} and \myblue{\textbf{lying}} to the public are the only possible strategies.\\
\midrule
\textbf{The Washington Times} (right): \mypink{\textit{Trump says he's ``doing very well''}} $\ldots$
\mypink{\textit{President Trump thanked the nation for supporting him}} Friday night as he left the White House to be hospitalized for COVID-19. \mypink{\textit{``I want to thank everybody for the tremendous support. $\ldots$''}} Mr. Trump said in a video recorded at the White House.\\
\midrule
\textbf{Breitbart} (right): \mypink{\textit{President Donald Trump thanked Americans for their support}} on Friday as he traveled to Walter Reed Military Hospital for further care after he was diagnosed with coronavirus. \mypink{\textit{``I think I'm doing very well$\ldots$"}} Trump said in a video filmed at the White House and posted to social media.\\
\bottomrule
\end{tabular}
\caption{
Article snippets by different media on the same news story. Contents that indicate stances and ideological leanings are highlighted in \myblue{\textbf{bold}} (for subjective phrases) and in \mypink{\textit{italics}} (for objective events). } 
\label{table-article-example}
\end{figure}

Using text as data, computational models for ideology measurement have rapidly expanded and diversified, including classical machine learning methods such as ideal point estimation~\cite{groseclose1999comparing,shor2011ideological}, Naive Bayes~\citep{Evans_McIntosh}, support vector machines~\citep{Yu_kaumann},  latent variable models~\cite{Barbera_jost}, and regression~\cite{peterson_spirling_2018}; and more recent neural architectures like recurrent neural networks~\cite{Iyyer_Enns} and Transformers~\cite{Baly_Martino,liu_aaai}.
Nonetheless, most of those models leverage datasets with ideology labels drawn from a single domain, and it is unclear if any of them can be generalized to diverse genres of text.

Trained on massive quantities of data, Pretrained Language Models (PLMs) have achieved state-of-the-art performance on many text classification problems, with an additional fine-tuning stage on labeled task-specific samples~\citep{bert,liu2019roberta}. 
Though PLMs suggest the promise of generalizable solutions, their ability to acquire the knowledge needed to detect complex features such as ideology from text across genres remains an open question.
PLMs have been shown to capture linguistic structures with a \textit{local focus}, such as task-specific words, syntactic agreement, and semantic compositionality~\cite{ClarkKLM19,JawaharSS19}. Although word choice is indicative of ideology, ideological leaning and stance are often revealed by which entities and events are selected for presentation~\cite{hackett1984decline,christie2005genre,enke2020you}, with the most notable strand of work in framing theory~\cite{entman1993,entman2007}. One such example is demonstrated in Figure~\ref{table-article-example}, where Daily Kos criticizes Trump's dishonesty while The Washington Times and Breitbart emphasize the good condition of his health.

In this work, we propose to train PLMs for a wide range of ideology-related downstream  tasks. We argue that it is critical for PLMs to consider the \textit{global context} of a given article. For instance, as pointed out by~\citet{basil}, one way to acquire such context is through comparison of news articles on the same story but reported by media of different ideologies.
Given the lack of suitable datasets, we first collect a new large-scale dataset, \textbf{\data}.\footnote{Our data and code can be accessed at \url{https://github.com/launchnlp/POLITICS}.} 
It contains $3{,}689{,}229$ English news articles on politics, gathered from 11 United States (US) media outlets covering a broad ideological spectrum. We further downsample
and  cluster articles in \data by different media into groups, each consisting of pieces aligned on the same story. The resultant dataset, \aligndata, contains $1{,}060{,}512$ stories with aligned articles.

Next we train a new PLM, \textbf{\model},
based on a \underline{P}retraining \underline{O}bjective \underline{L}everaging \underline{I}nter-article \underline{T}riplet-loss using \underline{I}deological \underline{C}ontent and \underline{S}tory. 
Concretely, we leverage continued pretraining~\cite{Gururangan2020continual}, where we design an \textbf{ideology objective} operating over clusters of \textit{same-story} articles to compact articles with similar ideology and contrast them with articles of different ideology. The learned representation can better discern the embedded ideological content. 
We further enhance it with a \textbf{story objective} that ensures the model to focus on meaningful content instead of overly relying on shortcuts, e.g., media boilerplate. Both objectives are used together with our specialized masked language model objective that focuses on entities and sentiments to train \model. 

Our main goal here is to create \textbf{general-purpose tools} for analyzing ideological content for researchers and practitioners in the \textbf{broad community}. 
Furthermore, when experimenting on 11 ideology prediction and stance detection tasks using 8 datasets of different genres, including a newly collected dataset from AllSides, \model outperforms both a strong SVM baseline and previous PLMs on 8 tasks. 
Notably, \model is particularly effective on long documents, e.g., achieving 10\% improvements on both ideology prediction and stance detection tasks over RoBERTa~\cite{liu2019roberta}. We further show that our model is more robust in setups with smaller training sets. 

\section{Related Work}
\textbf{Ideology prediction} is a critical task for quantitative political science~\citep{mullins_1972, Michael_Freeden, martin2015ideology, Wilkerson_Casas}. 
Both classical methods~\citep[e.g., Naive Bayes, SVM;][]{Evans_McIntosh,Yu_kaumann, Gheiler} and deep learning models~\citep[e.g., RNN;][]{Iyyer_Enns} have been used to predict ideology on a variety of datasets where ideology labels are available, such as legislative speeches \citep{laver_benoit_garry_2003} and U.S. Supreme Court briefs~\citep{Evans_McIntosh}.
Notably, \citet{liu_aaai} pretrains a Transformer-based language generator to minimize the ideological bias in generated text.
As generative models are not as effective as masked language models (MLMs) at text classification,
our goal differs in that we train MLMs to recognize ideological contents in various domains and tasks. 

\begin{table*}
\centering
\resizebox{0.92\linewidth}{!}{%
\begin{tabular}{ L{4em} R{4em} R{4em} R{4em} R{4em} R{4em} R{4em} R{4em} R{4em} R{4em} R{4em} R{4em} R{4em} }
\toprule
 & \textbf{Daily Kos} & \textbf{HPO} & \textbf{CNN} & \textbf{WaPo} & \textbf{NYT} & \textbf{USA Today} & \textbf{AP} & \textbf{The Hill} & \textbf{TWT} & \textbf{FOX} & \textbf{Breitbart} \\
\midrule
\textbf{Ideology} & L & L & L & L & L & C & C & C & R & R & R \\
\midrule
\textbf{\# articles} & 100,828 & 241,417 & 64,988 & 198,529 & 173,737 & 170,737 & 279,312 & 322,145 & 243,181 & 330,166 & 206,512 \\
\midrule
\textbf{\# words} & 738.7 & 729.9 & 655.7 & 803.2 & 599.4 & 691.7 & 572.3 & 426.3 & 522.7 & 773.5 & 483.5 \\
\bottomrule
\end{tabular}
}
\caption{
Statistics of \datad. Media outlets are sorted by ideology from left (L), center (C), to right (R) based on AllSides and Media Bias Chart. HPO: Huffington Post; WaPo: The Washington Post; NYT: The New York Times; 
TWT: The Washington Times.
Additional statistics of raw data size before downsampling and the corresponding publish dates can be found in Table \ref{table:append-corpus}.
}
\label{crawl-table}
\end{table*}

\smallskip
\noindent \textbf{Stance detection} is a useful task for ideology analysis because co-partisans are generally positive towards each other and negative towards counter-partisans~\cite{aref2021identifying}.
There has been a large body of work on identifying individuals' stances towards specific targets from the given text~\cite{ThomasPang,Walker_Anand,Hasan_Ng}.
On the methodology side, \newcite{Mohammad_Sobhani} and \newcite{stance_svm} apply statistical models, e.g., SVM, with handcrafted text features. Neural methods have also been widely investigated, including CNN~\citep{pkudblab}, LSTM~\citep{Augenstein_et_al}, hierarchical networks \citep{Sun_Wang}, and representation learning~\cite{Darwish_Stefanov}. 

Recent research focus resides in leveraging PLMs for predicting stances, e.g., incorporating extra features~\cite{Prakash_Madabushi}.
\citet{kawintiranon-singh-2021-knowledge} share a similar spirit with our work by upsampling tokens to mask. However, they pre-define a list of tokens customized for the given targets, which is hard to generalize to new targets. 
We aim to train PLMs \nop{with MLM objectives} relying on general-purpose sentiment lexicons and important entities, to foster model generalizability.

\smallskip
\noindent \textbf{Domain-specific Pretrained Language Models.} 
PLMs, such as BERT \citep{bert} and RoBERTa \citep{liu2019roberta}, have obtained state-of-the-art results on many NLP tasks. 
Inspired by the observation that a continued pretraining phase on in-domain data yields better  performance~\cite{Gururangan2020continual}, domain-specific PLMs are introduced~\cite{SciBERT,finbert,clinicalbert,biobert}. 
However, they only use the default MLM objective, without considering domain knowledge. 
In this work, we design ideology-driven pretraining objectives to inject domain knowledge to discern ideologies and related stances.

Focusing on the news domain, PLMs have been primarily used for factuality prediction~\cite{Jwa2019exBAKEAF,zellers2019grover,Kaliyar_Goswami} and topic classification~\cite{Liu_Xia, buyukoz-etal-2020-analyzing, guptapolibert} by fine-tuning on task-specific datasets. 
Few work has investigated PLMs for understanding political ideology evinced in texts. 
One exception is \citet{Baly_Martino}, where they also leverage the triplet loss as the pretraining objective. However, our work is novel in at least three aspects. 
First, our triplet loss is designed to capture the ideological (dis)similarity among articles \textit{on the same story}, while the loss used by \citet{Baly_Martino} operates on articles \textit{of the same topic}. As a result, their approach can falsely compact representations of very different news contents, e.g., articles on ``Japan Economics'' and ``Indian Troops'' both belong to the topic of ``Asia''. 
Moreover, our newly introduced story objective can effectively prevent the model from relying on media-specific language (e.g., ``for the New York Times"), while their objective may fail to do so, and thus lacks generalizability to languages used by different media and other ideology-related tasks. 
Finally, we use \data that contains more than $3$M articles, which is more suitable for pretraining large models than the small dataset ($35$k articles) used by \citet{Baly_Martino}. 
To the best of our knowledge, we are the first to systematically study and release PLMs for ideology-related study in the US political domain.

\section{Pretraining Datasets}
\label{section:pretraining_data}

\subsection{Data Crawling}
\label{section:data-crawling}
We collect pretraining datasets from online news articles with diverse ideological leanings and language usage. 
We select 11 media outlets based on their ideologies (from far-left to far-right)
and popularity.\footnote{We use \url{https://www.allsides.com} and \url{https://adfontesmedia.com} to decide ideology and \url{https://www.alexa.com/topsites} to decide popularity.} 
We convert their ideologies into three categories: left, center, and right,
and crawl all pages published by them between January 2000 and June 2021, from Common Crawl and Internet Archive.
We then follow \citet{raffel2020exploring} for data cleaning, and, additionally, only retain news articles related to US politics. 
Appendix \ref{appendix:data-clean} describes in detail the steps for removing non-articles pages, duplicates, non-US pages, and boilerplate. 

The cleaned data, dubbed \textbf{\data}, contains $3{,}689{,}229$ US political news articles. To mitigate the bias that some media dominate the model training, we downsample the corpus so that each ideology contributes equally. The downsampled corpus, \textbf{\datad}, contains $2{,}331{,}552$ news articles, with statistics listed in Table \ref{crawl-table}. We keep $30$K held-out articles as validation set.  
\nop{\datad is used to train all baselines and models in this work that employ a MLM objective. }

\subsection{Aligning Articles on the Same Story}
\label{section:alignment}
We compare how media outlets from different sides report the same story, which intuitively better captures ideological content. To this end, we design an algorithm to align articles in \datad that cover the same story. 
We treat each article as an anchor, and find matches from other outlets based on the following similarity score:

\vspace{-10pt}
{
\small
\begin{equation}
    \text{sim}(p_i, p_j)=\alpha * \text{sim}_{t}(p_i, p_j) + 
    (1-\alpha)* \text{sim}_{e}(p_i, p_j)
\label{equation:alignment}
\end{equation}
}

\noindent where $p_i$ and $p_j$ are two articles, $\text{sim}_{t}$ is the cosine similarity between TF-IDF vectors of $p_i$ and $p_j$, $\text{sim}_{e}$ is the weighted Jaccard similarity between the sets of named entities\footnote{Extracted by Stanford CoreNLP~\cite{manning-EtAl:2014:P14-5}.
}
in $p_i$ and $p_j$, and $\alpha=0.4$ is a hyperparameter. 
During alignment, for an article from an outlet to be considered as a match, it must be published within three days before or after the anchor, has the highest similarity score among articles from the same outlet, and the score is at least $\theta = 0.23$. 
Hyperparameters $\alpha$ and $\theta$ are searched on the Basil dataset~\citep{basil}, which contains manually aligned articles.\footnote{Our algorithm achieves a mean reciprocal rank of 0.612 on Basil, with detailed evaluation in Appendix \ref{appendix:alignment}.}
After deduplicating articles in each story cluster, we obtain \textbf{\aligndata}, containing $1{,}060{,}512$ clusters with an average of $4.29$ articles in each. 
Appendix \ref{appendix:alignment} details the alignment algorithm. 

\section{\model via Continued Pretraining}
\label{section:methodology}

Here we introduce our continued pretraining methods based on a newly proposed \textbf{ideology objective} that drives representation learning to better discern ideological content by comparing same-story articles (\cref{section:triplet-loss}), which is further augmented by a \textbf{story objective} to better focus on content. 
They are combined with the masked language model objective, which is tailored to focus on entities and sentiments (\cref{section:mlm}), to produce \model (\cref{section:objective}).

\subsection{Ideology-driven Pretraining Objectives}
\label{section:triplet-loss}

\begin{figure}[t]
    \centering
    \includegraphics[width=0.35\textwidth]{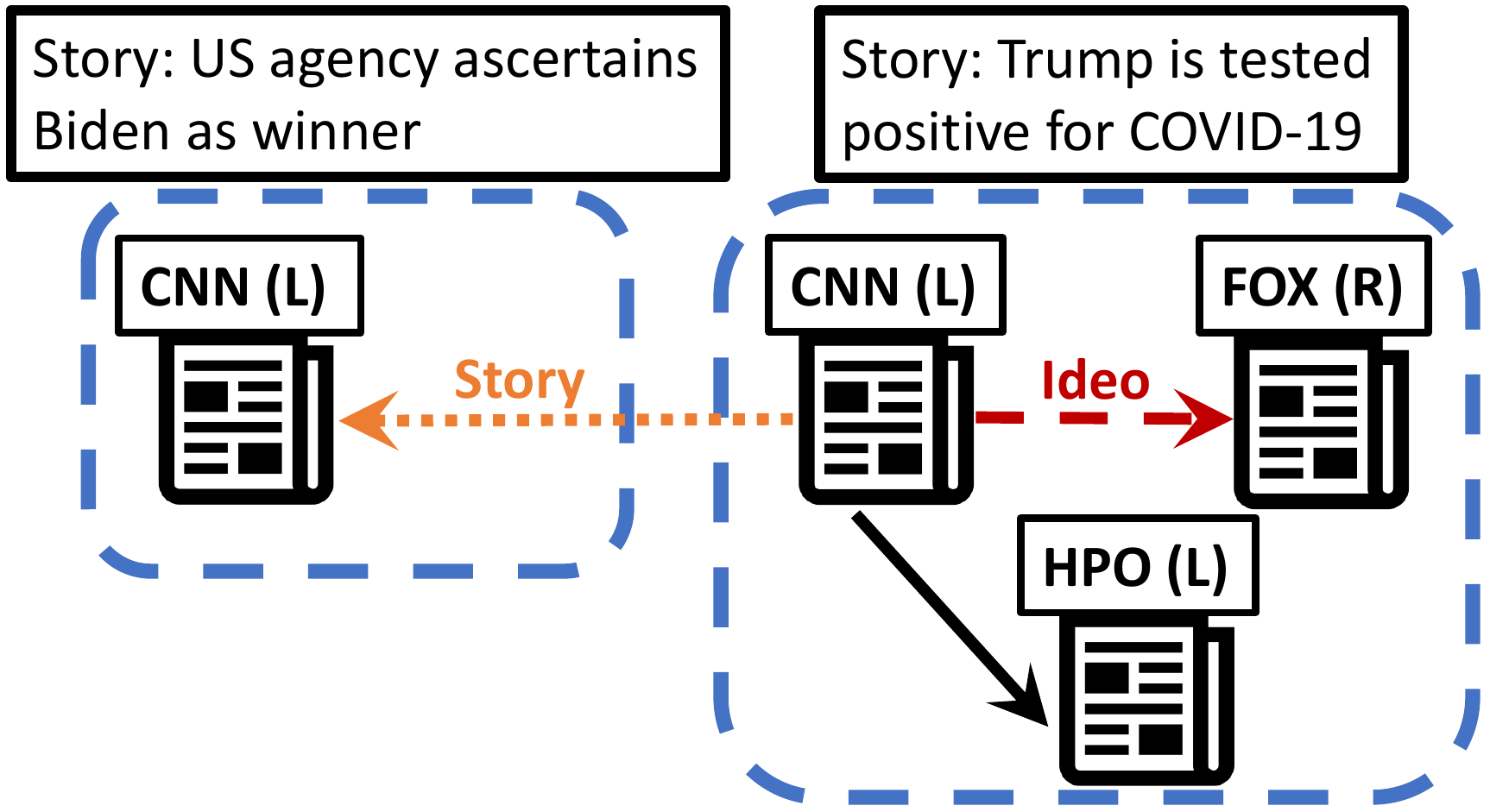}
    \caption{
    Construction of the ideology and story objectives. 
    The middle CNN article is the anchor in this example. 
    Solid black arrow represents positive-pair relation for both objectives; 
    \textcolor{myred}{red} dashed arrow denotes negative-pair for ideology objective; \textcolor{myorange}{orange} dashed arrow indicates negative-pair for story objective.
    }
    \label{fig:triplet}
\end{figure}

To promote representation learning that better captures ideological content, we leverage \aligndata with articles grouped by stories to provide story-level background for model training.
That is, we use triplet loss~\citep{2015facenet} that operates over \textbf{triplets} of $<$anchor, positive, negative$>$ to encourage anchor and positive samples to have closer representations while contrasting anchor from negative samples.

Our primary pretraining objective, i.e., {ideology objective}, uses the triplet loss to teach the model to acquire \textbf{ideology-informed representations} by comparing \textit{same-story} articles written by media of \textit{different ideologies}. As shown in Figure~\ref{fig:triplet}, given a story cluster, we choose an article published by media on the left or right as the \textit{anchor}. We then take articles in the same cluster with the same ideology as \textit{positive} samples, and articles with the opposite ideology as \textit{negative} ones. The ideology objective is formulated as follows: 

\vspace{-5pt}
{
\small
\begin{equation}
    \mathcal{L}_{\text{ideo}}=\sum_{t\in\mathcal{T}_{\text{ideo}}}\left[\norm{\vectorbold{t^{(a)}}-\vectorbold{t^{(p)}}}_2 - \norm{\vectorbold{t^{(a)}}-\vectorbold{t^{(n)}}}_2 + \delta_{\text{ideo}}\right]_+ \label{eq:ideo_loss}
\end{equation}
}

\noindent where $\mathcal{T}_{\text{ideo}}$ is the set of all ideology triplets, $\vectorbold{t^{(a)}}$, $\vectorbold{t^{(p)}}$, and $\vectorbold{t^{(n)}}$ are the \texttt{[CLS]} representations of anchor, positive, and negative articles in triplet $t$, $\delta_{\text{ideo}}$ is a hyperparameter, and $\left[\cdot \right]_+$ is  $max(\cdot, 0)$.

Next, we augment the ideology objective with a {story objective} to allow the model to focus on \textbf{semantically meaningful content} and to prevent the model from focusing on ``shortcuts'' (such as media-specific languages) to detect ideology.
To construct story triplets,
we use the same $<$anchor, positive$>$ pairs as in the ideology triplet, and then treat articles from the same media outlet but on different stories as negative samples, as depicted in Figure~\ref{fig:triplet}. 
Similarly, our story objective is formulated as follows: 

\vspace{-5pt}
{
\small
\begin{equation}
   \mathcal{L}_{\text{story}}=\sum_{t\in\mathcal{T}_{\text{story}}}\left[\norm{\vectorbold{t^{(a)}}-\vectorbold{t^{(p)}}}_2 - \norm{\vectorbold{t^{(a)}}-\vectorbold{t^{(n)}}}_2 + \delta_{\text{story}}\right]_+ \label{eq:story_loss}
\end{equation}
}%

\noindent where $\mathcal{T}_{\text{story}}$ contains all story triplets, and $\delta_{\text{story}}$ is a hyperparameter searched on the validation set.

\subsection{Entity- and Sentiment-aware MLM}
\label{section:mlm}

Here we present a specialized MLM objective to collaborate with our triplet loss based objectives for better representation learning. Notably, political framing effect is often reflected in which entities are selected for reporting~\cite{Gentzkow_Shapiro}. Moreover, the occurrence of sentimental content along with the entities also signal stances~\cite{Mohammad_Sobhani}. 
Therefore, we take a masking strategy that upsamples \textit{entity} tokens~\cite{sun2019ernie, guu2020realm, kawintiranon-singh-2021-knowledge} and \textit{sentiment} words to be masked for the MLM objective, which improves from prior pretraining work that only considers article-level comparison~\cite{Baly_Martino}.

Concretely, we consider named entities with types of PERSON, NORP, ORG, GPE and EVENT. We detect sentiment words using lexicons by~\citet{10.1145/1014052.1014073} and~\citet{wilson-etal-2005-recognizing}. 
To allow MLM training to focus on entities and sentiment, we mask them with a $30\%$ probability, and then randomly mask remaining tokens until $15\%$ of all tokens are reached, as done in \newcite{bert}. Masked tokens are replaced with \texttt{[MASK]}, random tokens, and original tokens with a ratio of 8:1:1. 

\subsection{Overall Pretraining Objective}
\label{section:objective}
We combine the aforementioned objectives as our final pretraining objective as follows:

\vspace{-10pt}
{\fontsize{10}{11}\selectfont
\begin{equation}
    \mathcal{L}= \beta * \mathcal{L}_{\text{ideology}} + \gamma * \mathcal{L}_{\text{story}} + (1-\beta-\gamma) * \mathcal{L}_{\text{MLM}}
\end{equation}
}

\noindent where $\beta=\gamma=0.25$. Using $\mathcal{L}$, \model is produced via continued training on RoBERTa~\citep{liu2019roberta}.\footnote{We use \texttt{roberta-base} model card from Huggingface.} 
We do not try to train the model from scratch since \datad only has $\sim$10GB data, smaller than corpus for RoBERTa ($\sim$160GB). 
Hyperparameters are listed in Table \ref{table:append-pretrain}.

\section{Experiments}
\label{section:experiments}
\begin{table}
\centering
\resizebox{1.0\linewidth}{!}{%
\begin{tabular}{lrrrrrr}
\toprule
\textbf{Data} & \textbf{Genre} & \textbf{\# Train} & \textbf{Len.}  & \textbf{Split} \\
\midrule
Congress Speech~\cite{congress-speech} & speech & 7,000 & 538 & rand. \\
\midrule
AllSides (\textit{newly collected}) & news & 7,878 & 863 & time \\
\midrule
BASIL-article~\cite{basil} & news & 450 & 693 & story \\
\midrule
BASIL-sentence~\cite{basil} & news & 1,197 & 27 & story \\
\midrule
Hyperpartisan~\cite{kiesel-etal-2019-semeval} & news & 425 & 556 & rand. \\
\midrule
VAST~\cite{allaway-mckeown-2020-zero} & cmt & 11,545 & 102 & rand.$^\dagger$ \\
\midrule
YouTube User~\cite{wu2021crosspartisan} & cmt & 1,114 & 1,213 & user \\
\midrule
YouTube Cmt~\cite{wu2021crosspartisan} & cmt & 6,832 & 197 & user \\
\midrule
SemEval~\cite{semeval} & tweet & 2,251 & 17 &  rand.$^\dagger$ \\
\midrule
Twitter~\cite{preotiuc-pietro-etal-2017-beyond} & tweet & 1,079 & 2,298 &  user \\
\bottomrule
\end{tabular}
}
\caption{
Datasets used for evaluating PLMs vary in text genre, training set size (\# Train), length (Len.), and split criterion. 
Rand. denotes random split. 
Time split means training on the ``past'' data and test on the ``future''. 
Story split divides articles according to story clusters. 
User split indicates users in the test are unseen in the training. 
$^\dagger$: by the original work.
}
\label{table:benchmark}
\end{table}

\begin{table*}
\centering

\resizebox{1.0\linewidth}{!}{%
\begin{tabular}{lcccccccccccccc}
\toprule
\multicolumn{1}{c}{\multirow{3}{*}{\textbf{}}} & \multicolumn{7}{c}{\textbf{Ideology Prediction}} & \multicolumn{5}{c}{\textbf{Stance Detection}} & \multicolumn{1}{c}{} & \multicolumn{1}{c}{\multirow{3}{*}{\pbox{3em}{All \\ avg}}} \\
\cmidrule(lr){2-8}
\cmidrule(lr){9-14}

\multicolumn{1}{c}{} & \multicolumn{1}{c}{YT} & \multicolumn{1}{c}{\multirow{2}{*}{CongS}} & \multicolumn{1}{c}{\multirow{2}{*}{HP}} & \multicolumn{1}{c}{\multirow{2}{*}{AllS}} & \multicolumn{1}{c}{YT} & \multicolumn{1}{c}{\multirow{2}{*}{TW}} & \multicolumn{1}{c}{Ideo.}  & \multicolumn{1}{c}{SEval} & \multicolumn{1}{c}{SEval} & \multicolumn{1}{c}{Basil} & \multicolumn{1}{c}{\multirow{2}{*}{VAST}} & \multicolumn{1}{c}{Basil} & \multicolumn{1}{c}{Stan.}  & \multicolumn{1}{c}{} \\
\multicolumn{1}{c}{} & \multicolumn{1}{c}{(cmt.)} & \multicolumn{1}{c}{} & \multicolumn{1}{c}{} & \multicolumn{1}{c}{} & \multicolumn{1}{c}{(user)} & \multicolumn{1}{c}{} & \multicolumn{1}{c}{avg} & \multicolumn{1}{c}{(seen)} & \multicolumn{1}{c}{(unseen)} & \multicolumn{1}{c}{(sent.)} & \multicolumn{1}{c}{} & \multicolumn{1}{c}{(art.)} & \multicolumn{1}{c}{avg} & \multicolumn{1}{c}{} \\
\midrule
\multicolumn{15}{l}{\textbf{Baselines}} \\
SVM & 65.34 & {\ul 71.31} & 61.25 & 52.51 & 66.49 & 42.85 & 59.96 & 51.18 & 32.89 & 51.08 & 39.54 & 30.77 & 41.09 & 51.38 \\
BERT & 64.64 & 65.88 & 48.42 & 60.88 & 65.24 & 44.20 & 58.21 & 65.07 & 40.39 & 62.81 & 70.53 & 45.61 & 56.88 & 57.61 \\
RoBERTa & 66.72 & 67.25 & 60.43 & 74.75 & 67.98 & 48.90 & 64.34 & \textbf{70.15} & 63.08 & 68.16 & 76.25 & 41.36 & 63.80 & 64.09 \\
\midrule
\multicolumn{15}{l}{\textbf{\newcite{Baly_Martino}}} \\
with Original Data & 65.42 & 66.74 & 58.37 & 72.89 & 70.47 & 44.95 & 63.14 & 68.66 & 56.29 & 61.30 & 75.57 & 37.98 & 59.96 & 61.69 \\
with \datad & 68.57 & 70.39 & \textbf{71.24} & {\ul 76.47} & 74.74 & 47.38 & {\ul 68.13} & 65.84 & 49.54 & 60.60 & 75.03 & 41.84 & 58.57 & 63.79 \\

\midrule
\multicolumn{15}{l}{\textbf{Our models with triplet loss objective only}} \\

Ideology Obj. & 66.20 & \colorbox{cyan!15}{68.18} & \colorbox{cyan!15}{ 64.15} & \colorbox{cyan!15}{\textbf{76.52}} & \colorbox{cyan!15}{68.15} & 42.66 & 64.31 & 68.78 & 59.61 & 64.18 & 76.03 & \colorbox{cyan!15}{44.94} & 62.71 & 63.58 \\
Story Obj. & 66.09 & \colorbox{cyan!15}{69.11} & 56.70 & 74.59 & \colorbox{cyan!15}{68.89} & 46.53 & 63.65 & 69.02 & \colorbox{cyan!15}{\textbf{63.54}} & 67.21 & \colorbox{cyan!15}{76.66} & \colorbox{cyan!15}{\textbf{53.16}} & \colorbox{cyan!15}{{\ul 65.92}} & \colorbox{cyan!15}{64.68} \\
Ideology Obj. + Story Obj. & \colorbox{cyan!15}{{\ul 68.91}} & \colorbox{cyan!15}{69.10} & \colorbox{cyan!15}{63.08} & \colorbox{cyan!15}{ 76.23} & \colorbox{cyan!15}{{\ul 77.58}} & \colorbox{cyan!15}{\textbf{48.98}} & \colorbox{cyan!15}{ 67.31} & 69.66 & \colorbox{cyan!15}{{\ul 63.17}} & 64.37 & 76.18 & \colorbox{cyan!15}{47.01} & \colorbox{cyan!15}{64.08} & \colorbox{cyan!15}{{\ul 65.84}} \\

\noalign{\vskip 0.3ex}
\hdashline[5pt/4pt]
\noalign{\vskip 0.3ex}
\multicolumn{15}{l}{\textbf{Our models with masked language model objective only}} \\
Random & 67.82 & 70.32 & 60.59 & 73.54 & 70.77 & 44.62 & 64.61 & 69.16 & 60.39 & 69.94 & \textbf{77.11} & 39.16 & 63.15 & 63.95 \\
Upsamp. Ent. & \colorbox{green!15}{\textbf{69.06}} & \colorbox{green!15}{70.32} & 60.09 & 70.89 & \colorbox{green!15}{71.40} & \colorbox{green!15}{47.16} & \colorbox{green!15}{64.82} & \colorbox{green!15}{69.81} & \colorbox{green!15}{63.08} & 69.49 & 76.76 & \colorbox{green!15}{46.46} & \colorbox{green!15}{65.12} & \colorbox{green!15}{64.96} \\
Upsamp. Sentiment & 67.41 & 70.03 & 56.05 & 72.35 & \colorbox{green!15}{74.93} & \colorbox{green!15}{48.15} & \colorbox{green!15}{64.82} & \colorbox{green!15}{{\ul 70.09}} & \colorbox{green!15}{60.81} &  \colorbox{green!15}{{\ul 71.28}} & 76.61 & \colorbox{green!15}{44.42} & \colorbox{green!15}{64.64} & \colorbox{green!15}{64.74} \\
Upsamp. Ent. + Sentiment & \colorbox{green!15}{68.31} & \colorbox{green!15}{\textbf{71.42}} & 58.02 & 71.90 & \colorbox{green!15}{71.04} & \colorbox{green!15}{47.31} & \colorbox{green!15}{64.67} & \colorbox{green!15}{69.25} & \colorbox{green!15}{62.84} & 69.23 & {\ul 77.10} & \colorbox{green!15}{43.16} & \colorbox{green!15}{64.32} & \colorbox{green!15}{64.51} \\
\noalign{\vskip 0.3ex}
\hdashline[5pt/4pt]
\noalign{\vskip 0.3ex}
\model & \colorbox{magenta!15}{67.83$^\ast$} & 70.86 & \colorbox{magenta!15}{{\ul 70.25}$^\ast$} & \colorbox{magenta!15}{74.93} & \colorbox{magenta!15}{\textbf{78.73}$^\ast$} & \colorbox{magenta!15}{{\ul 48.92}} & \colorbox{magenta!15}{\textbf{68.59}} & 69.41 & 61.26 & \colorbox{magenta!15}{\textbf{73.41$^\ast$}} & \colorbox{magenta!15}{76.73$^\ast$} & \colorbox{magenta!15}{{\ul 51.94}$^\ast$} & \colorbox{magenta!15}{\textbf{66.55}} & \colorbox{magenta!15}{\textbf{67.66}} \\
\bottomrule
\end{tabular}
}
\caption{
Macro F1 scores on 11 evaluation tasks (average of 5 runs). 
Tasks are sorted by text length, short to long, within each group. 
``All avg'' is the average of all 11 tasks. 
\textbf{Best} results are in bold and \underline{second best} are underlined. 
Our models with triplet-loss objectives that outperform RoBERTa are in \colorbox{cyan!15}{blue}. 
Our models with specialized sampling methods that outperform the one with vanilla MLM (Random) are in \colorbox{green!15}{green}. 
\model uses Ideology + Story Obj. and Upsamp. Ent. + Sentiment. 
Results where \model outperforms all baselines are in \colorbox{magenta!15}{red}, with $^\ast$ indicating statistical significance \citep[Mann–Whitney U test; ][$p \leq 0.05$]{mann-whitney_test}. 
 Standard deviations (std) are reported in Table~\ref{table:results_new}.
The range of std over tasks is $[0.31, 3.42]$ for \model, and $[0.48, 7.35]$ for RoBERTa. 
}
\label{table:results}
\end{table*}

Given the importance of ideology prediction and stance detection tasks in political science  \cite{ThomasPang, Wilkerson_Casas, Chatsiou_Mikhaylov}, we conduct extensive experiments on a wide spectrum of datasets with 11 tasks (\cref{section:benchmarks}). We then compare with both classical models and prior PLMs (\cref{section:baselines}), and among our model variants (\cref{section:variants}). We present and discuss results in \cref{section:results}, where \model outperform all three baselines on 8 out of 11 tasks. 
For all models, MLM objectives are trained with \datad, and ideology and story objectives are trained on \aligndata. Details are in Appendix~\ref{app:continuted_pretraining}.

\subsection{Datasets and Tasks}
\label{section:benchmarks}

Our tasks are discussed below,
with statistics listed in Table \ref{table:benchmark} and more descriptions in Appendix~\ref{appendix:dataset}. 

\smallskip
\noindent\textbf{Ideology prediction} tasks for predicting the political leanings are evaluated on the following datasets. 

\begin{itemize}[leftmargin=1em,noitemsep,topsep=0pt,parsep=0pt,partopsep=0pt]
    \item \texttt{Congress Speech} \citep[\textbf{CongS};][]{congress-speech} contains speeches from US congressional records, each labeled as liberal or conservative. 
    
    \item \texttt{AllSides}
    \footnote{\url{https://www.allsides.com}.} 
    (\textbf{AllS}, new) is a website that assesses political bias and ideology of US media. In this study, we collect articles from AllSides with their ideological leanings on a 5-point scale. 
    
    \item \texttt{Hyperpartisan} \citep[\textbf{HP};][]{kiesel-etal-2019-semeval} is a shared task of predicting a binary label for an article as being hyperpartisan or not. We convert it into a 3-way classification task by splitting hyperpartisan news into left and right. 
    
    \item \texttt{YouTube} \citep[\textbf{YT;}][]{wu2021crosspartisan} contains discussions on YouTube. \textbf{cmt.} and \textbf{user} refer to predicting left/right at the comment- and user-level, respectively.
    
    \item \texttt{Twitter} \citep[\textbf{TW;}][]{preotiuc-pietro-etal-2017-beyond}
    collects a group of Twitter users with self-reported ideologies on a 7-point scale. We merge them into 3-way labels. \nop{Each user is represented by their posted tweets.}
\end{itemize}

\smallskip
\noindent\textbf{Stance detection} tasks, which predict a subject's attitude towards a given target from a piece of text, are listed below. All tasks take a 3-way label (positive, negative, and neutral) except for \textbf{BASIL (sent.)} that labels positive or negative. 

\begin{itemize}[leftmargin=1em,noitemsep,topsep=0pt,parsep=0pt,partopsep=0pt]
    \item \texttt{BASIL} \citep{basil} contains news articles with annotations on authors' stances towards entities. \textbf{BASIL (sent.)} and \textbf{BASIL (art.)} are prediction tasks at sentence and article-levels.
    
    \item \texttt{VAST} \citep{allaway-mckeown-2020-zero} collects online comments from \nop{The New York Times ``Room for Debate'' section} {``Room for Debate''}, with stances labeled towards the debate topic. 
    
    \item \texttt{SemEval} \cite{semeval} is a shared task on detecting stances in tweets. We consider two setups to predict on seen, i.e. \textbf{SEval (seen)}, and unseen, i.e., \textbf{SEval (unseen)}, entities. 
    
\end{itemize}

\subsection{Baselines}
\label{section:baselines}

We consider three baselines. First, we train a linear \textbf{SVM} using unigram and bigram features for each task, since it is a common baseline in political science~\citep{Yu_kaumann,diermeier2012language}. Hyperparameters and feature selection are described in Table \ref{table:append-svm}. 
We further compare with \textbf{BERT} and \textbf{RoBERTa}, following the standard fine-tuning process for ideology prediction tasks and using the prompt described in \cref{section:fine-tune} for stance detection.

\subsection{Model Variants}
\label{section:variants}
We consider several variants of \model. First, using \textbf{triplet loss objective only}, we experiment on models trained  with ideology objective (\textit{Ideology Obj.}), story objective (\textit{Story Obj.}), or both. 

Next, we continue pretaining RoBERTa with \textbf{MLM objective only}, using vanilla MLM objective (\textit{Random}), entity focused objective (\textit{Upsamp. Ent.}), sentiment focused objective (\textit{Upsamp. Sentiment}), or upsampling both entity and sentiment.

\subsection{Fine-tuning Procedure}
\label{section:fine-tune}

We fine-tune each neural model for up to 10 epochs, with early stopping enabled. We select the best fine-tuned model on validation sets using F1. Details of experimental setups are in Table \ref{table:append-fine-tune}. 

\paragraph{Ideology Prediction.}
We follow common practice of using the \texttt{[CLS]} token for standard fine-tuning \citep{bert}. 
For \texttt{Twitter} and \texttt{YouTube User} data, we encode them using sliding windows and aggregate by mean pooling.

\paragraph{Stance Detection.} 
We follow \citet{schick2021exploiting} on using prompts to fine-tune models for stance detection. We curate 11 prompts (in Table \ref{table:append-prompt}) and choose the best one based on the average F1 by RoBERTa on all stance detection tasks:

\vspace{-20pt}

$$p \texttt{[SEP]}\text{\textit{The stance towards} \{target\} \textit{is} \texttt{[MASK]} .}$$

\vspace{-5pt}

\noindent The model is trained to predict \texttt{[MASK]} for stance, conditioned on the input $p$ and \{target\}.

\subsection{Main Results}
\label{section:results}

Table~\ref{table:results} presents F1 scores on all tasks. 
\model achieves the best overall average F1 score across the board, 3.6\% better than the strongest baseline, RoBERTa. 
More importantly, \model alone outperforms all three baselines listed in \cref{section:baselines} on 8 out of 11 tasks,
including more than 10\% of improvement for ideology labeling on \texttt{Hyperpartisan} and \texttt{Youtube} user-level. 
We attribute the performance gain to our proposed ideology-driven pretraining objective, which helps capture partisan content. 
Note that, on some tasks, other model variants lead \model by a small margin, and this may be of interest to practitioners performing specific tasks. 

We further compare with the model proposed by \citet{Baly_Martino}, which also leverages triplet loss as pretraining objective but on articles of the same topics. 
We implement two versions of their model, using the original data released by \citet{Baly_Martino}\footnote{\url{https://github.com/ramybaly/Article-Bias-Prediction}.} and our \datad. 
First, pretraining on our \datad yields better results on ideology prediction tasks than using the original data, indicating the value of \datad. Second, using the triple construction method by \citet{Baly_Martino} with \datad does not generalize well on the stance detection task, compared to \model and its variants. This highlights the advantage of our objectives that enable content comparison among articles of the same stories. 

Moreover, \textit{our ideology-driven objectives help acquire knowledge needed to discern ideology as well as stance detection.} 
When equipping the RoBERTa model with ideology and story objectives but no MLM objective, it achieves the second best overall performance.

Next, \textit{focusing on entities better identifies stance}. Simply continuing training RoBERTa with vanilla MLM objective (\textit{Random}) does not yield performance gain on stance detection, while our upsampling methods make a difference, i.e., increasing sampling ratios of entities improves F1 by 2\%. 

\begin{table}
\centering
\resizebox{1.0\linewidth}{!}{%
\begin{tabular}{lcccc}
\toprule
\multicolumn{1}{c}{\multirow{2}{*}{Models}} & \multicolumn{4}{c}{Hyperpartisan} \\
\cmidrule(lr){2-5}
\multicolumn{1}{c}{} & Acc. & Precision & Recall & F1 \\
\midrule
\citealp{JiangPSBM19} (ELMo+CNN)& 82.2 & 87.1 & 75.5 & 80.9 \\
\citealp{SrivastavaGPSRK19} (Logistic Regression)& 82.0 & 81.5 & 82.8 & 82.1 \\
\citealp{HanawaSOSI19} (BERT) & 80.9 & 82.3 & 78.7 & 80.5 \\
\citealp{IsbisterJ19} (SVM) & 80.6 & 85.8 & 73.2 & 79.0 \\
\citealp{YehLS19} (ULMFiT) & 80.3 & 79.3 & 81.8 & 80.6 \\
\midrule
RoBERTa & 84.3 & \textbf{87.2} & 80.6 & 83.7 \\
\model & \textbf{85.2} & 86.3 & \textbf{83.7} & \textbf{84.9}  \\
\bottomrule
\end{tabular}
}
\caption{
Comparison with the previous state-of-the-art models on \texttt{Hyperpartisan} using the original binary prediction setup. Best results are in bold. \model obtains the best accuracy, recall and F1. Note, precision, recall, and F1 are measured for the hyperpartisan class.}
\label{table:sota_hp}
\end{table}
\begin{table*}
\centering
\resizebox{1.0\linewidth}{!}{%
\begin{tabular}{lcccccccccccccc}
\toprule
\multicolumn{1}{c}{\multirow{3}{*}{\textbf{}}} & \multicolumn{7}{c}{\textbf{Ideology Prediction}} & \multicolumn{5}{c}{\textbf{Stance Detection}} & \multicolumn{1}{c}{} & \multicolumn{1}{c}{\multirow{3}{*}{\pbox{3em}{All \\ avg}}} \\
\cmidrule(lr){2-8}
\cmidrule(lr){9-14}

\multicolumn{1}{c}{} & \multicolumn{1}{c}{YT} &\multicolumn{1}{c}{\multirow{2}{*}{CongS}} & \multicolumn{1}{c}{\multirow{2}{*}{HP}} & \multicolumn{1}{c}{\multirow{2}{*}{AllS}} & \multicolumn{1}{c}{YT} & \multicolumn{1}{c}{\multirow{2}{*}{TW}} & \multicolumn{1}{c}{Ideo.}  & \multicolumn{1}{c}{SEval} & \multicolumn{1}{c}{SEval} & \multicolumn{1}{c}{Basil} & \multicolumn{1}{c}{\multirow{2}{*}{VAST}} & \multicolumn{1}{c}{Basil} & \multicolumn{1}{c}{Stan.}  & \multicolumn{1}{c}{} \\
\multicolumn{1}{c}{} & \multicolumn{1}{c}{(cmt.)} & \multicolumn{1}{c}{} & \multicolumn{1}{c}{} & \multicolumn{1}{c}{} & \multicolumn{1}{c}{(user)} & \multicolumn{1}{c}{} & \multicolumn{1}{c}{avg} & \multicolumn{1}{c}{(seen)} & \multicolumn{1}{c}{(unseen)} & \multicolumn{1}{c}{(sent.)} & \multicolumn{1}{c}{} & \multicolumn{1}{c}{(art.)} & \multicolumn{1}{c}{avg} & \multicolumn{1}{c}{} \\
\midrule

\model & 67.83 & 70.86 & 70.25 & 74.93 & 78.73 & \textbf{48.92} & 68.59 & 69.41 & 61.26 & \textbf{73.41} & 76.73 & 51.94 & \textbf{66.55} & \textbf{67.66} \\
\midrule
No Ideology Obj. & \colorbox{cyan!50}{  -3.78} & \colorbox{cyan!15}{  -2.17} &  \colorbox{cyan!50}{  -16.35} & \colorbox{cyan!50}{  -3.28} &  \colorbox{cyan!50}{  -12.54} & \colorbox{cyan!50}{ -3.43} &  \colorbox{cyan!50}{  -6.93} & -0.38 & {  -0.83} &  \colorbox{cyan!50}{  -4.22} & -0.45 &  \colorbox{cyan!50}{  -16.01} &  \colorbox{cyan!50}{  -4.38} &  \colorbox{cyan!50}{  -5.77} \\
No Story Obj. &  \colorbox{magenta!15}{\textbf{+1.98}} & +0.64 & -0.72 & +0.70 & +0.29 &  \colorbox{cyan!15}{-1.78} & +0.19 &  \colorbox{cyan!15}{  -1.23} &  \colorbox{magenta!50}{\textbf{+2.94}} &  \colorbox{cyan!50}{-3.36} & -0.87 &  \colorbox{cyan!50}{-10.75} &  \colorbox{cyan!50}{-2.66} &  \colorbox{cyan!15}{-1.11} \\
No Upsamp. Ent. & +0.18 & -0.65 & -0.05 & +0.55 & -0.29 &  \colorbox{cyan!15}{-1.20} & -0.24 & \textbf{+0.62} & -0.67 & \ \colorbox{cyan!50}{-3.74} & -0.55 &  \colorbox{cyan!15}{-1.20} &  \colorbox{cyan!15}{-1.11} & -0.64 \\
No Upsamp. Sentiment & +0.75 & -0.28 & \textbf{+0.22} &  \colorbox{cyan!15}{-1.27 }& -0.11 &  \colorbox{cyan!15}{-1.40} & -0.35 & -0.84 &  \colorbox{magenta!15}{+1.67} &  \colorbox{cyan!50}{-3.91} &  \colorbox{cyan!15}{  -1.10} &  \colorbox{magenta!15}{\textbf{+1.44}} & -0.55 & -0.44 \\
\model + Ideo. Pred. &  \colorbox{magenta!15}{+1.46} &  \colorbox{magenta!15}{\textbf{+1.10}} &  \colorbox{cyan!15}{-1.01} &  \colorbox{magenta!50}{\textbf{+4.72}} & \colorbox{magenta!50}{\textbf{+2.02}} &  \colorbox{cyan!50}{  -3.96} & \textbf{+0.72} & +0.41 & -0.52 & \colorbox{cyan!50}{-3.82} & \textbf{+0.12} & \colorbox{cyan!50}{-3.10} & \colorbox{cyan!15}{-1.38} & -0.23 \\

\bottomrule
\end{tabular}
}
\caption{
 Ablation study results on \model. 
 \model + Ideo. Pred.: triplet-loss objective is replaced with a hard label prediction objective on ideology of articles (left vs. right). \textbf{Best} results are in bold.
 Darker \colorbox{magenta!50}{red} shows greater improvements. 
 Darker \colorbox{cyan!50}{blue} indicates larger performance drop. 
 The ideology objective contributes the most to \model, followed by the story objective. 
}
\label{table:abltation}
\end{table*}

\smallskip
\noindent \textbf{Comparisons with Previous State-of-the-arts.} 
Using the \textit{original} binary prediction setup (i.e., hyperpartisan or not) on \texttt{Hyperpartisan} data~\cite{kiesel-etal-2019-semeval}, \model obtains an accuracy of $85.2$, leading previous state-of-the-art results by at least $3$ points, as shown in Table~\ref{table:sota_hp}.

POLITICS achieves an F1 of $77.0$ on the \textit{original} \texttt{VAST} data where the previous state-of-the-art model obtained $69.2$~\cite{Jayaram_Allaway}. 
On \texttt{SemEval}, \model yields an F1 of $71.3$ where the best performance is $76.5$ by \newcite{Ghadir_Azmi}.
Notably, we adopt one single classifier in our setup, while they include separate models for different prediction targets, which have been shown to outperform one single classifier~\cite{semeval}.  
Full comparisons with previous methods are included in Appendix~\ref{app:SOTA_comparison}. 
For other tasks, there is no direct comparison as the datasets are either originally used for different prediction tasks (e.g., Basil is used for detecting media bias spans) or newly collected by this work.

\begin{figure}
    \centering
    \includegraphics[width=0.45\textwidth]{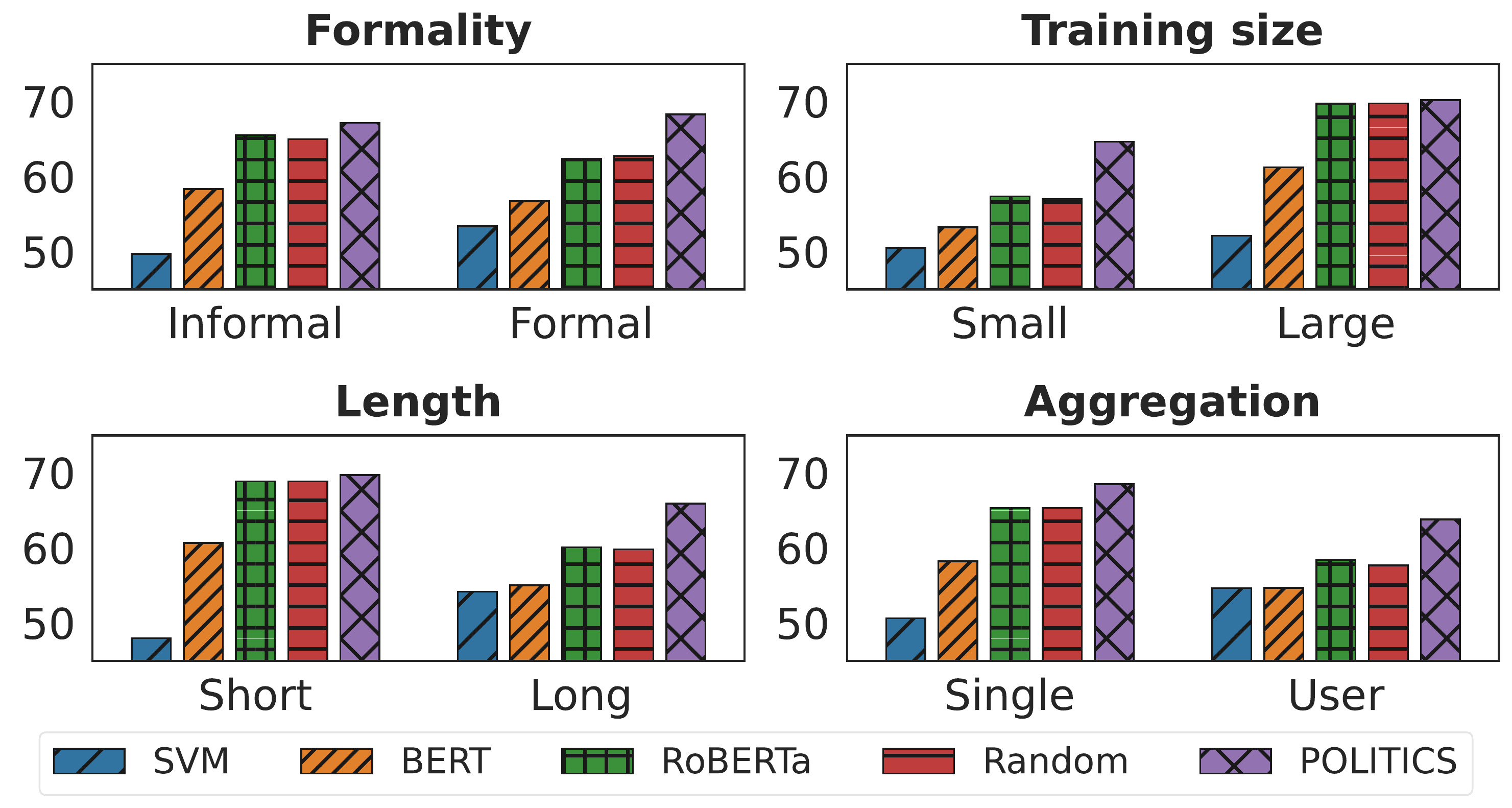}
    \caption{
    Macro F1 aggregated over tasks of different \textit{formality}, \textit{training size}, \textit{document length} and \textit{aggregation method (single post vs. posts by each user)}. 
    \model performs better on handling formal language, small training sets, and longer text. 
    }
    \label{fig:grouping}
\end{figure}

\smallskip
\noindent \textbf{On Texts of Different Characteristics.} 
Based on Table~\ref{table:benchmark}, we further study the model's performance on data of different properties: 
\textit{language formality}, \textit{training size}, \textit{document length}, and \textit{aggregation level}.
As shown in Figure~\ref{fig:grouping}, with each property (concrete criterion in Appendix \ref{appendix:property}), we divide tasks into two categories. 
\model yields greater improvements on more formal and longer text, since pretraining is done on news articles. \model is also more robust to training sets with small sizes, showing the potential effectiveness in few-shot learning, which is echoed in \cref{section:few-shot}.

\section{Further Analyses}

\nop{\sout{In this section, we first explore the few-shot ability of \model as it is critical for applications in political science. \add{For example, some politicians or journalists are often unwilling to disclose ideology.}
We then conduct ablation studies and remove one component of \model at a time to see the contribution of each component to the final performance gain.
We also visualize \model to examine what features or knowledge it captures. Lastly, we measure the bias encoded in \model to see model's fairness and robustness.
}}

\subsection{Few-shot Learning}
\label{section:few-shot}

\begin{figure}[t]
    \centering
    \includegraphics[width=0.48\textwidth]{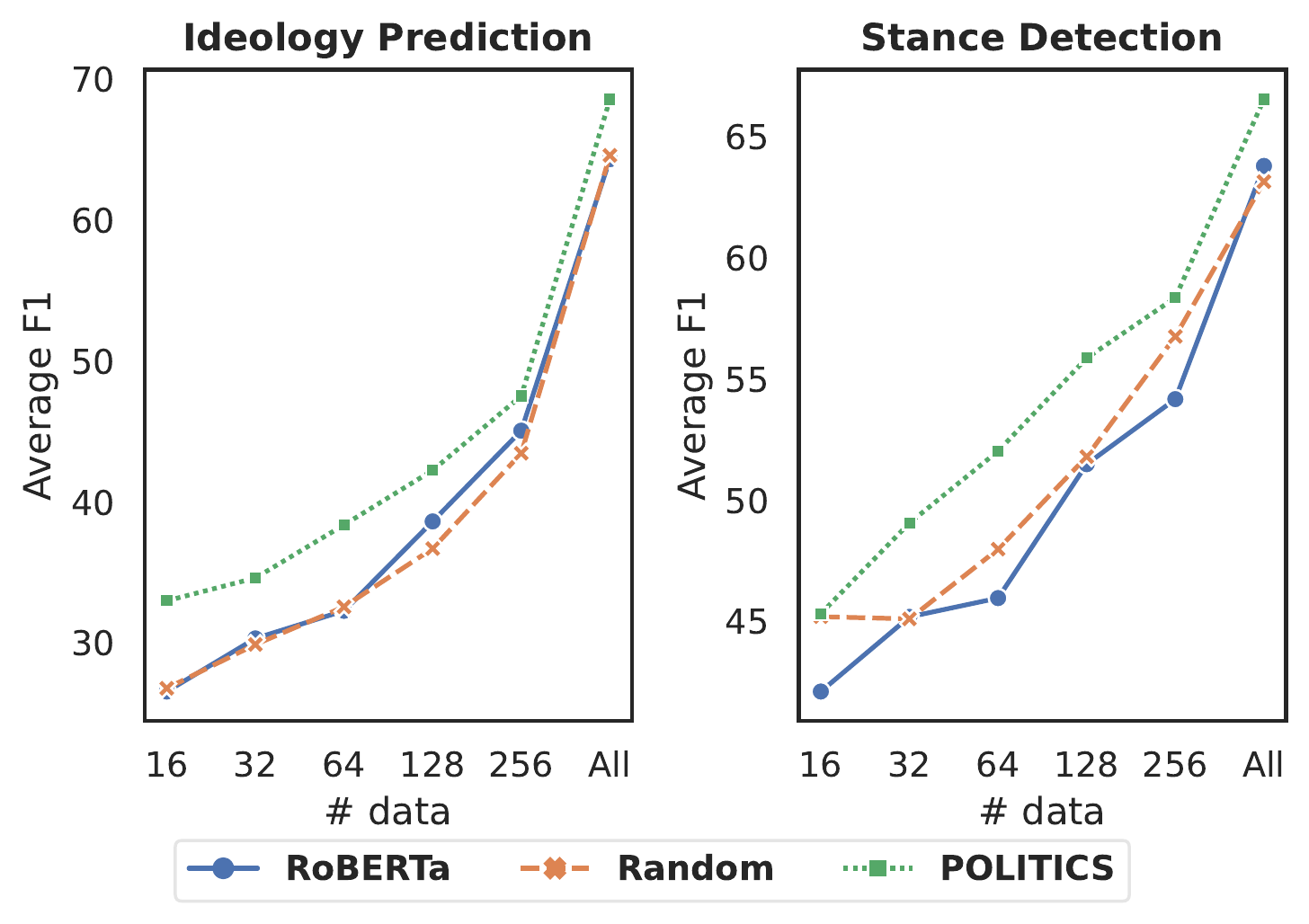}
    \caption{
    Average of ideology prediction and stance detection performances with few-shot learning. \model uniformly outperforms RoBERTa and it being continued pretrained with vanilla MLM (\textit{Random}). 
    }
    \label{fig:few-shot}
\end{figure}

We first fine-tune all PLMs on small numbers of samples. 
\model consistently outperforms the two counterparts on both tasks, using small training sets (Figure~\ref{fig:few-shot}).
More importantly, naively training RoBERTa on the large \datad does not help ideology prediction. 
By contrast, our ideology-driven objective can better capture ideology than the baselines, even when using only 16 samples for fine-tuning on the ideology tasks.

\begin{figure*}
    \centering
    \includegraphics[width=\textwidth]{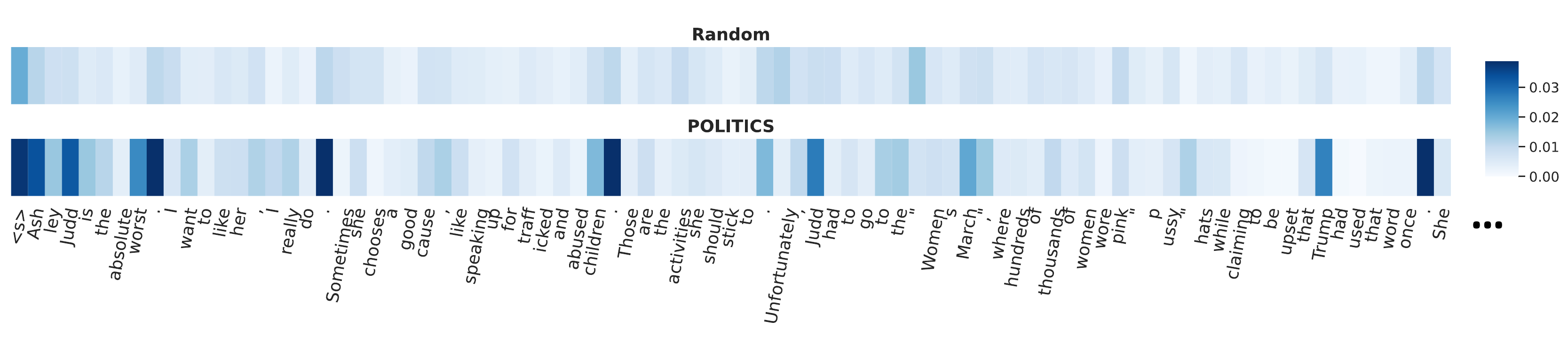}
    \caption{
    Last layer attention scores between \texttt{[CLS]} token and other input tokens (aggregated over all heads).
    \model captures ``Ashley Judd'', ``worst'', and ``Trump''. 
    }
    \label{fig:attention-short}
\end{figure*}

\subsection{Ablation Study on \model}

We show the impact of removing each ideology-driven pretraining objective and upsampling strategy from \model in Table~\ref{table:abltation}. 
First, removing the ideology objective results in the most loss on both tasks. This again demonstrates the effectiveness of our triplet-loss formulation over same-story articles. 
Removing the story objective also hurts the overall performance by $1\%$ but improves the ideology prediction marginally. This shows that the story objective functions as an auxiliary constraint to avoid over-fitting on the ``shortcuts'' for discerning ideologies. 
Moreover, removing upsampling strategies generally weakens \model's performance, but only to a limited extent.

We also experiment with a setup with hard-ideology learning (i.e., directly predicting the ideology of each article without using triplet-loss objectives). Not surprisingly, this variant (\textit{\model+Ideo. Pred.}) outperforms \model on ideology prediction since it can directly learn ideology from the annotated labels. However, it has been overfitted to ideology prediction tasks and lacks generalizability, thus yields worse performance on stance detection.

\subsection{Visualizing Attentions}
On the \texttt{Hyperpartisan} task, we visualize the last layer's attention weights between the \texttt{[CLS]} token and all other tokens by \model and RoBERTa pretrained with vanilla MLM on \datad (\textit{Random}). 
We randomly sample 20 test articles, and for 13 of them, \model is able to capture salient entities, events, and sentiments in the text whereas \textit{Random} cannot. We present one example in Figure~\ref{fig:attention-short} where \model captures ``Ashley Judd'', ``the worst'', and ``Trump''. More examples are given in Appendix \ref{section:appendix-attention-viz}. 
This finding confirms that our ideology-driven objective and upsampling strategies can help the model focus more on entities of political interest as well as better recognize sentiments.

\begin{figure}[t]
    \centering
    \includegraphics[width=0.38\textwidth]{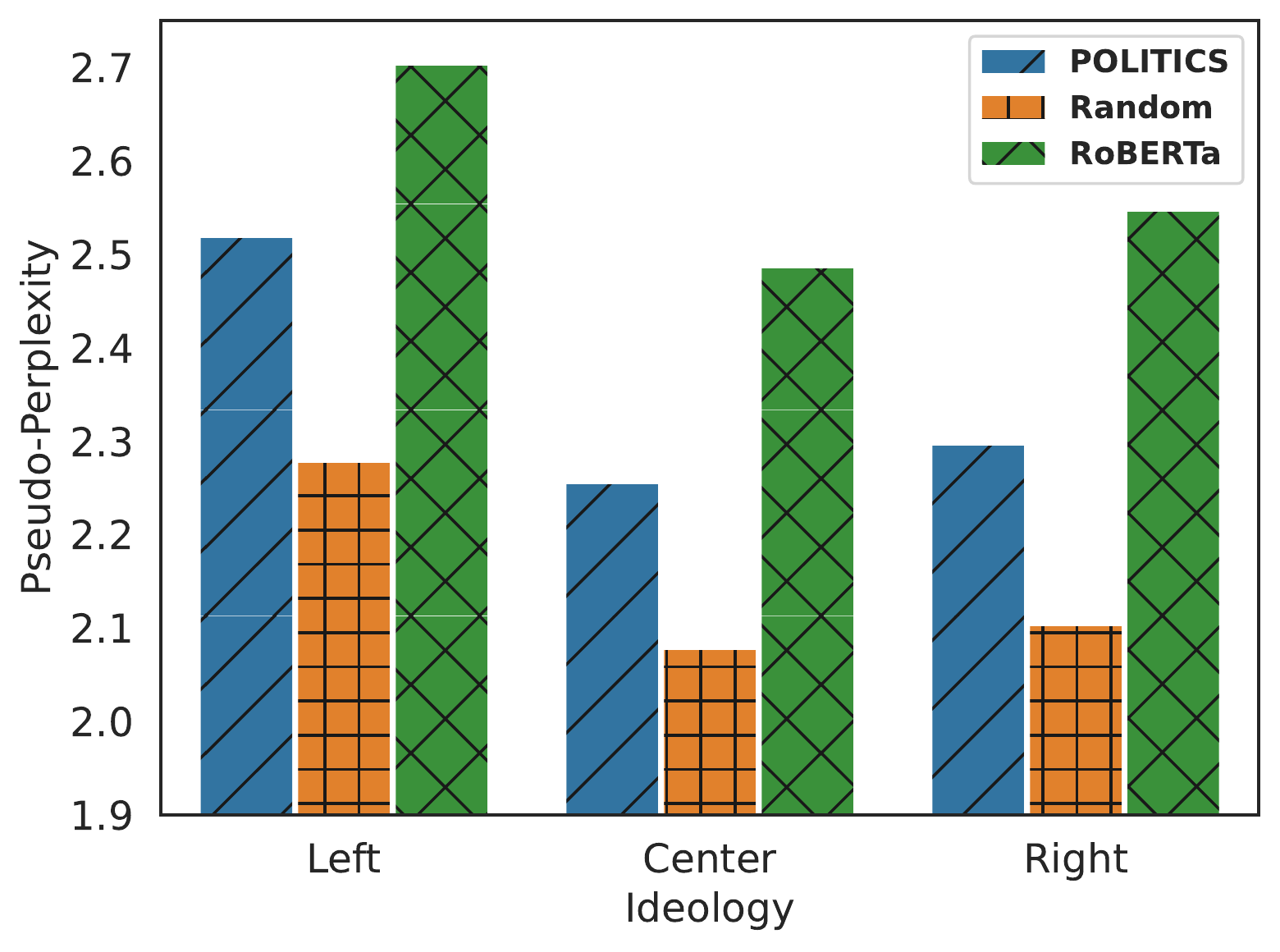}
    \caption{
    Model perplexities  on 30K validation articles in \datad. Perplexities do not drop sharply on \model compared with RoBERTa being continued pretrained with MLM objective (\textit{Random}), suggesting \model can yield superior predictive performance while not overfitting with ideological languages. 
    }
    \label{fig:ppl}
    \vspace{-5pt}
\end{figure}

\subsection{\model on Different Ideologies}

Finally, we measure whether PLMs would acquire ideological bias as measured by whether they fit with languages used by a specific ideology. 
Concretely, we follow~\citet{salazar-etal-2020-masked} to evaluate PLMs on $30$K held-out articles of different ideologies from \datad with \textit{pseudo-perplexity}. For efficiency, we estimate the \textit{pseudo log-likelihood} based on 200 random tokens in each article as used by~\citet{wang-cho-2019-bert}. 
As illustrated in Figure \ref{fig:ppl}, while MLM objective (\textit{Random}) is effective at fitting a corpus, i.e., having the lowest perplexities, triplet-loss objectives act as regularizers during pretraining, shown by the higher perplexity of \model compared to \textit{Random}. 
Interestingly, we find center and right articles have lower perplexity than that of left articles. We hypothesize that it relates to political science findings that, over recent periods of political polarization in US, Republicans have become somewhat more coherent and similar than Democrats \citep{grossmann2016asymmetric,benkler2018network}, and are thus easier to predict.

\section{Conclusion}

We study the problem of training general-purpose tools for ideology content understanding and prediction. 
We present \model, trained with novel ideology-driven pretraining objectives based on the comparisons of same-story articles written by media outlets of different ideologies. To facilitate model training, we also collect a large-scale dataset, \data, consisting of news articles of different ideological leanings. 
Experiments on diverse datasets for ideology prediction and stance detection tasks show that \model outperforms strong baselines, even with a limited amount of labeled samples for training, and state-of-the-art models.

\section*{Acknowledgments}
This work is supported in part through National Science Foundation under grants IIS-2100885 and IIS-2127747, and computational resources and services provided by Advanced Research Computing (ARC), a division of Information and Technology Services (ITS) at the University of Michigan, Ann Arbor. 
We appreciate the anonymous reviewers for their helpful comments. 
We thank the members of the LAUNCH group at the University of Michigan for discussions and suggestions. We also thank Changyuan Qiu for helping collect the AllSides articles, Siqi Wu for giving us access to the Youtube comments, and Daniel Preotiuc-Pietro for sharing the Twitter user accounts used in the original study to allow the collection of the corresponding users' tweets in this work.

\section{Ethical Considerations}
\subsection{\data Collection}
All news articles were collected in a manner consistent with the terms of use of the original sources as well as the intellectual property and the privacy rights of the original authors of the texts, i.e., source owners. During data collection, the authors honored privacy rights of content creators, thus did not collect any sensitive information that can reveal their identities. All participants involved in the process have completed human subjects research training at their affiliated institutions. We also consulted Section 107\footnote{\url{https://www.copyright.gov/title17/92chap1.html\#107}.} of the U.S. Copyright Act and ensured that our collection action fell under the fair use category.

\subsection{Dataset Usage}
All of the newly collected datasets in this work will be made available upon request. 
Pretraining corpus details are included in Section~\ref{section:pretraining_data}.  
The other seven datasets used for downstream evaluation are obtained in the following ways.  \textbf{CongS}, \textbf{HP}, \textbf{BASIL}, \textbf{VAST} and \textbf{SEval} are acquired by direct download. 
\textbf{CongS} is released under the ODC-BY 1.0 license (free to share, create, and adapt). 
\textbf{HP} and \textbf{SEval} are developed in shared tasks by the NLP community, which allow the use of copyrighted material without permission from the copyright holder for research purposes \citep{EscartinRLMWL17}. 
For \textbf{VAST}, the author explicitly states ``We make our dataset and models available for use''. 
\textbf{BASIL} is developed by the last author and her collaborators. 
For \textbf{YT} and \textbf{TW}, we consult with the corresponding authors and obtain the datasets by agreeing that we will not further distribute them. 
Dataset details are listed in Section~\ref{section:benchmarks} and Appendix~\ref{appendix:dataset}.

\subsection{Benefit and Potential Misuse}

\noindent \textbf{Intended use.} The models developed in this work can assist the general public to measure and understand ideological language used in diverse genres of texts. For example, \model can help the general public know where their representatives stand on key issues. Our experiments in Section~\ref{section:experiments} demonstrate how \model would be deployed in real life when handling applications in both ideology prediction and stance detection. We deem that our extensive experiments have covered the major usage of \model.

\noindent \textbf{Failure mode} is defined as situations where \model fails to correctly predict the ideology of an individual or a given text. 
In such cases, \model might deliver misinformation or cause misunderstanding towards a political figure or a policy. For vulnerable populations (e.g., people who maybe not be able to make the right judgements), the harm could be tremendously magnified when they fail to interpret the model outputs or blindly trust machine responses. 
Ideally, the interpretation of our model's predictions should be carried out within the broader context of the source text. 

\smallskip
\noindent \textbf{Misuse potential.}  Users may mistakenly take the machine prediction as a golden rule or a fact.
We would recommend any politics-related machine learning models, including ours, put up an ``use with caution'' message to encourage users to check more sources or consult political science experts to reduce the risk of being misled by single source. 
Moreover, \model might also be misused to label people with a specific political leaning that they do not want to be associated with. We suggest that when in use the tools should be accompanied with descriptions about their limitations and imperfect performance, as well as allow users to opt out from being the subjects of measurement. 

\smallskip
\noindent \textbf{Potential limitation.} Although multiple genres are considered, the genre coverage is not exhaustive, and does not include other trending media or content of different modalities for expressing opinions, such as TV transcripts, images, and videos. Thus, the predictive performance of \model may still be under investigated. Further, in downstream evaluation, \model is only trained and tested in the same domain, so its cross-genre ability needs further evaluation.

\smallskip
\noindent \textbf{Bias Mitigation.} During data preprocessing, we create \datad to ensure that all ideologies have almost equal presence to minimize potential bias.
\model is not designed to encode bias. In Figure~\ref{fig:ppl}, the discrepancy in perplexities among different ideologies is more related to the greater coherence among Republicans than Democrats,
rather than \model encoding biased knowledge.

In conclusion, there is no greater than minimal risk/harm introduced by either \datad or \model. However, to discourage the misuse, we will always warn users that model predictions are for informational purpose only and users should always resort to the broader context to reduce the risk of absorbing biased information.

\clearpage


\bibliographystyle{acl_natbib}

\clearpage

\setcounter{table}{0}
\setcounter{figure}{0}
\renewcommand{\thefigure}{A\arabic{figure}}
\renewcommand{\thetable}{A\arabic{table}}

\begin{appendices}
\section{\data Cleaning Steps}
\label{appendix:data-clean}
In this section, we provide the details of our data cleaning steps for \data. We adopt the following cleaning steps to only keep news articles that relate to US politics.

\paragraph{Removing Non-article Pages.}
Online news websites also post non-news content. We remove such pages by checking their page titles and URLs based on a list of patterns. Sample patterns are shown in Table \ref{non arti-table}.

\paragraph {Removing Duplicate Pages.}
We use character-level edit distance to identify duplicate pages. Specifically, we use the following formula to calculate the difference between page $a$ and page $b$:
\vspace{-10pt}
\begin{equation}
    \text{diff}(a,b)=\text{dist}(a,b)/\max(\text{len}(a), \text{len}(b))
\end{equation}
where $\text{dist}(a, b)$ is the Levenshtein distance between $a$ and $b$. If the value is less than 0.1, we consider two pages as duplicates and we only keep the one with earlier publication date. Following this procedure, we remove duplicated pages within each media outlet.

\begin{table}[h]
\centering
\resizebox{0.95\linewidth}{!}{%
\begin{tabular}{ll}
\toprule  & \textbf{Filter Patterns} \\ \midrule
URL & /video/, /gallery/, /slideshow/ \\
\midrule
\multirow{3}{2em}{Title} & weekly digest, 10 sites you should know, \\
& day's end roundup, photos of the week,\\
& 5 things you need to know \\
\bottomrule
\end{tabular}
}
\caption{\label{non arti-table} Examples of patterns used to filter out pages that are not news stories.}
\end{table}

\paragraph {Removing Non-politics Pages.}
To filter out non-politics pages, we build a classifier using training data from \data. Since URL typically indicates a page's topic, we use keywords in the URL to retrieve politics and non-politics training data. The lists of keywords are shown in Table \ref{trn data-table}. This results in a training dataset with $400,462$ politics pages and $310,377$ non-politics pages. 
We also randomly sample $888$ pages from the remaining dataset and manually annotate them to use as the test set.

\begin{table}
\centering
\resizebox{0.95\linewidth}{!}{%
\begin{tabular}{ll}
\toprule   & \textbf{Keywords} \\ \midrule
\multirow{2}{4em}{Politics} & /politics/, /political/, /policy/, \\
& /election/, /elections/, /allpolitics/ \\
\midrule
\multirow{4}{4em}{Non-politics} & /travel/, /sports/, /life/, /movie/, \\
& /entertainment/, /science/, /music/, \\
& /plated/, /leisure/, /showbiz/, \\
& /lifestyle/, /fashion/, /art/, /sport/ \\
\bottomrule
\end{tabular}
}
\caption{Full list of keywords used to retrieve positive and negative training data for the politics classifier.}
\label{trn data-table}
\end{table}

We train a logistic regression model based on unigram and bigram TF-IDF features. 
To include pages not covered by the lists of keywords in Table \ref{trn data-table}, we use the trained classifier to classify remaining pages and add those classified with high probability\footnote{We use 0.95 for politics pages and 0.9 for non-politics pages.} 
to the training data. This results in a larger training set with 957,424 politics pages and 987,898 non-politics pages. We train the final classifier on the larger training set, and achieve an 
$88.18\%$ accuracy on the test data.

\paragraph {Removing Non-US Pages.}
We filter out pages that are not related to US by searching for non-US keywords in the URL.
For each of those pages, we only remove it if its text contains no US-related keywords.
Examples of keywords used are shown in Table \ref{non us-table}.

\begin{table}
\centering
\resizebox{0.95\linewidth}{!}{%
\begin{tabular}{ll}
\toprule \textbf{URL Keywords} & \textbf{Text Keywords} \\ \midrule
\makecell{/world/, /international/,\\ /europe/, /africa/, \\ /asia/, /latin-america/,\\ /middle-east/} & \makecell{U.S., United States,\\ Obama, Trump, Bush, \\Biden, Pompeo, \\ Clinton, Pence} \\
\bottomrule
\end{tabular}
}
\caption{Examples of keywords used to filter out non-US articles. For text keywords, we collect the names of all presidents, vice presidents, and secretaries of state of US since 2000.}
\label{non us-table}
\end{table}

\begin{table*}
\centering
\resizebox{0.8\linewidth}{!}{%
\begin{tabular}{lccc}
\toprule
& \textbf{\# article before downsample} & \textbf{Earliest date} & \textbf{Latest date} \\ \midrule
Daily Kos                  & 235,244                             & 2009-01-02                & 2021-06-30              \\ \midrule
HuffPost (HPO)             & 560,581                             & 2000-11-30                & 2021-06-30              \\ \midrule
CNN                        & 152,579                             & 2000-01-01                & 2021-06-30              \\ \midrule
The Washington Post (WaPo) & 461,032                             & 2000-01-01                & 2021-06-30              \\ \midrule
The New York Times (NYT)   & 403,191                             & 2000-01-01                & 2021-06-22              \\ \midrule
USA Today                  & 174,525                             & 2001-01-01                & 2021-06-30              \\ \midrule
Associated Press (AP)      & 285,685                             & 2000-01-01                & 2021-06-30              \\ \midrule
The Hill (Hill)            & 337,256                             & 2002-10-06                & 2021-06-30              \\ \midrule
The Washington Times (TWT) & 336,056                             & 2000-01-01                & 2021-06-30              \\ \midrule
Fox News (FOX)             & 457,550                             & 2001-01-12                & 2021-06-25              \\ \midrule
Breitbart News (Breitbart) & 285,530                             & 2009-01-08                & 2021-06-30              \\ \bottomrule
\end{tabular}
}
\caption{Statistics of \data corpus. Media outlets are sorted by ideology from left to right.}
\label{table:append-corpus}
\end{table*}

\paragraph{Removing Media Leaking Phrases.}
To prevent the model from learning features specific to individual media outlets, we perform a two-step cleaning.
First, we mask phrases that mention the media outlet itself (e.g., New York Times, NYTimes, and nytimes.com). Second, we create a list of patterns for frequently appearing sentences (more than 100 times), for each media outlet.
For example, as in ``\texttt{author} currently serves as a senior political analyst for \texttt{[MASK]} Channel and contributes to all major political coverage'', both the author name and the sentence itself can leak media outlet information. 
Since sentences with media leaking information usually appear at the beginning or end of the article, we remove any of the first and last two paragraphs, if they contain a sentence that matches such pattern.

\begin{table}
\centering
\resizebox{0.85\linewidth}{!}{%
\begin{tabular}{l p{8em}}
\toprule
\textbf{Hyperparameter} & \textbf{Value}           \\ \midrule
number of steps         & 2,500                    \\ \midrule
batch size              & 2048                     \\ \midrule
maximum learning rate   & 0.0005                   \\ \midrule
learning rate scheduler & linear decay with warmup \\ \midrule
warmup percentage       & 6\%                      \\ \midrule
optimizer               & AdamW \citep{adamw}                  \\ \midrule
weight decay            & 0.01                     \\ \midrule
AdamW beta weights      & 0.9, 0.98
\\ \midrule
$\delta_{\text{ideo}}$      & 0.5
\\ \midrule
$\delta_{\text{story}}$      & 1.0
\\ \bottomrule
\end{tabular}
}
\caption{Hyperparameters for continued pretraining.}
\label{table:append-pretrain}
\end{table}

\section{News Story Alignment}
\label{appendix:alignment}
As shown in Equation \ref{equation:alignment}, we combine text similarity and entity similarity to be the final story similarity score. Only title and the first five sentences are considered in the calculation. We further require aligned articles $a$ and $b$ to satisfy two constraints:
\begin{itemize}[leftmargin=1em,noitemsep,topsep=2pt,parsep=0pt,partopsep=2pt]
    \item The difference in publication dates of $a$ and $b$ is at most three days.
    \item $a$ and $b$ must contain at least one common named entity in the title or in the first three sentences.
\end{itemize}
We use CoreNLP to extract named entities in articles \cite{manning-EtAl:2014:P14-5}. For the second constraint, we further apply Crosswikis to map each entity to a unique concept in Wikipedia \cite{spitkovsky-chang-2012-cross}. When calculating entity similarity, we split each entity into single words and remove stop words. After alignment, we use the procedure described in Appendix \ref{appendix:data-clean} to remove duplicate articles in the same story cluster. The hyperparameters are $\alpha=0.4$ and $\theta=0.23$.

\paragraph {Evaluating Alignment Algorithm.}
We search the hyperparameters on the Basil dataset \citep{basil} and test the algorithm on the Allsides dataset collected in \citet{cao-wang-2021-inference}. The Allsides dataset consists of manually aligned news articles from $251$ media outlets. After removing media outlets not in \datad, we obtain $2,904$ articles on $1,316$ stories.

To evaluate the performance of the alignment algorithm, we add the evaluation dataset into \datad and treat each evaluation article as the anchor article for the alignment algorithm. We use the remaining evaluation articles in the same story as relevant articles, which becomes the target to be identified. The algorithm achieves $0.612$ mean reciprocal rank (MRR) on the Basil dataset and $0.679$ MRR on the Allsides dataset.

\begin{table*}
\centering
\resizebox{0.85\linewidth}{!}{%
\begin{tabular}{ll}
\toprule
\textbf{Prompt} & \textbf{Verbalizer} \\
\midrule
$p$ \texttt{[SEP]} \textit{The stance towards} \{target\} \textit{is} \texttt{[MASK]}. & negative or positive\\
$p$ \texttt{[SEP]} \textit{It reveals a} \texttt{[MASK]} \textit{stance on} \{target\}. & negative or positive\\
$p$ \texttt{[SEP]} \textit{The speaker holds a} \texttt{[MASK]} \textit{attitude towards} \{target\}. & negative or positive\\
$p$ \texttt{[SEP]} \textit{What is the stance on} \{target\}? \texttt{[MASK]}. & Negative or Positive\\
$p$ \texttt{[SEP]} \textit{The previous passage} \texttt{[MASK]} \{target\}. & opposes or favors\\
$p$ \texttt{[SEP]} \textit{The stance on} \{target\} \textit{is} \texttt{[MASK]}. & negative or positive\\
$p$ \texttt{[SEP]} \textit{The stance towards} \{target\}: \texttt{[MASK]}. & negative or positive\\
$p$ \texttt{[SEP]} \textit{The author} \texttt{[MASK]} \{target\}. & opposes or favors\\
$p$ \texttt{[SEP]}\texttt{[MASK]} \{target\} & oppose or favor\\
$p$ \texttt{[SEP]}\texttt{[MASK]}. \{target\} & No or Yes \\
$p$ \texttt{[SEP]}\texttt{[MASK]} \{target\} & No or Yes \\
\bottomrule
\end{tabular}
}
\caption{List of prompts used for stance detection tasks. $p$ is the input text, and \{target\} is the target of interest. Verbalizer maps the label (e.g., against) to the token (e.g., negative). Some datasets use a third label (neutral). }
\label{table:append-prompt}
\end{table*}

\section{Continued Pretraining and Fine-tuning}
\subsection{Continued Pretraining}
\label{app:continuted_pretraining}
We initialize all variants of \model with a RoBERTa-base model \citep{liu2019roberta}, which contains about 125M parameters. Our implementation is based on the HuggingFace transformers library \citep{wolf-etal-2020-transformers}.\footnote{\url{https://github.com/huggingface/transformers}.}
\begin{table}
\centering
\resizebox{0.85\linewidth}{!}{%
\begin{tabular}{l p{7em}}
\toprule
\textbf{Hyperparameter} & \textbf{Value}           \\ \midrule
number of epochs        & 10                       \\ \midrule
patience                & 4                        \\ \midrule
maximum learning rate   & 0.00001 or 0.00002       \\ \midrule
learning rate scheduler & linear decay with warmup \\ \midrule
warmup percentage       & 6\%                      \\ \midrule
optimizer               & AdamW                    \\ \midrule
weight decay            & 0.001                    \\ \midrule
AdamW beta weights      & 0.9, 0.999               \\ \midrule
\# FFNN layer   & 2 \\
\midrule
hidden layer dimension in FFNN & 768 \\
\midrule
dropout in FFNN          & 0.1 \\
\midrule
sliding window size     & 512                      \\ \midrule
sliding window overlap  & 64 \\
\bottomrule
\end{tabular}
}
\caption{Hyperparameters used to fine-tune PLMs.}
\label{table:append-fine-tune}
\vspace{-5pt}
\end{table}
We train each model using 8 Quadro RTX 8000 GPUs for $2,500$ steps.
The total training time for \model is 20 hours, with shorter time for other variants.
Table \ref{table:append-pretrain} lists the training hyperparameters.

\paragraph{Training Details.}
For triplet loss objectives, we only consider triplets in each mini-batch. We skip a batch if it contains no triplet.
For the MLM objective, we truncate the article if it has more than 512 tokens. When masking entities and sentiment words, we only consider those with at most five tokens.
When both triplet loss and MLM objectives are enabled, we adopt alternating training strategy as in \citet{Ganin_Ustinova} to apply these two objectives for parameter updates in an alternating manner.

\begin{table}
\centering
\resizebox{0.85\linewidth}{!}{%
\begin{tabular}{l p{6em}}
\toprule
\textbf{Hyperparameter} & \textbf{Value}           \\ \midrule
kernel              & linear                   \\ \midrule
regularization strength     & 0.3, 1, or 3     \\
\midrule
features              & unigram and bigram TF-IDF                   \\ \midrule
minimum document frequency              & 5                   \\ \midrule
maximum document frequency    &  $0.7*|D|$ \\
\bottomrule
\end{tabular}
}
\caption{Hyperparameters used to train SVM. $|D|$ is the number of documents in the training set.}
\label{table:append-svm}
\end{table}

\subsection{Fine-tuning}
For both ideology prediction and stance detection tasks, we fine-tune each model for up to 10 epochs. We use early stopping and select the best checkpoint on validation set among 10 epochs. For ideology prediction tasks, we follow standard practice of using \texttt{[CLS]} token and feed-forward neural networks (FFNN) for classification. For stance detection tasks, we use prompts to fine-tune PLMs. We curate 11 prompts as shown in Table \ref{table:append-prompt}, and select the best prompt based on the performance of RoBERTa. Fine-tuning hyperparameters are listed in Table \ref{table:append-fine-tune}. 

For the SVM classifier, we use the implementation of TF-IDF feature extractor and linear SVM classifier in scikit-learn \citep{scikit-learn}. The classifier's hyperparameters are listed in Table \ref{table:append-svm}.

\begin{table}
\centering
\resizebox{1.0\linewidth}{!}{%
\begin{tabular}{lccc}
\toprule
\multicolumn{1}{c}{\multirow{2}{*}{Models}}               & \multicolumn{3}{c}{VAST}                      \\
               \cmidrule(lr){2-4}
\multicolumn{1}{c}{}                 & $\text{F}_{favor}$         & $\text{F}_{against}$       & $\text{F}_{avg}$          \\
                \midrule
BERT-joint \citep{allaway-mckeown-2020-zero}     & 54.5          & 59.1          & 65.3          \\
TGA Net \citep{allaway-mckeown-2020-zero}        & 57.3          & 59.0            & 66.5          \\
BERT-base \citep{Jayaram_Allaway}           & 64.3          & 58.1          & 69.2          \\
prior-bin:gold \citep{Jayaram_Allaway} & 64.5 & 54.6          & 68.4          \\
\midrule
RoBERTa       & 67.2          & 71.4          & 76.5 \\
\model         & \textbf{68.0}          & \textbf{72.2} & \textbf{77.0}          \\
\bottomrule
\end{tabular}
}
\caption{Comparison with state-of-the-art results on the original VAST dataset. Best results are in bold. }
\label{table:sota_vast}
\end{table}

\section{Downstream Evaluation Datasets}
\label{appendix:dataset}
This section lists more details of the eight datasets used in our downstream evaluation as well as their processing steps.

\subsection{Ideology Prediction}
\begin{itemize}[leftmargin=1em,noitemsep,topsep=0pt,parsep=0pt,partopsep=0pt]
    \item \texttt{Congress Speech}\footnote{\url{https://data.stanford.edu/congress\_text}.} \citep[\textbf{CongS};][]{congress-speech}:  We filter out speeches with less than 80 words and use the speaker's party affiliation as the ideology of the speech.
    
    \item \texttt{AllSides}\footnote{\url{https://www.allsides.com}.} (\textbf{AllS}): We crawl articles from AllSides and use the media outlet's annotated ideology as that of the article.
    
    \item \texttt{Hyperpartisan}\footnote{\url{https://webis.de/data/pan-semeval-hyperpartisan-news-detection-19.html}.} \citep[\textbf{HP};][]{kiesel-etal-2019-semeval}:  We convert the benchmark into a 3-way classification task by projecting media-level ideology annotations to articles.

    \item \texttt{YouTube} \citep{wu2021crosspartisan} contains cross-partisan discussions between liberals and conservatives on YouTube. In our experiments. we only keep controversial comments: 1) A video must have at least 1,500 comments and 150,000 views; 2) A comment must have at least 20 replies. The original dataset annotates users' ideology on a 7-point scale. We further convert it into a 3-way classification task for left, center, and right ideologies. For the comment-level prediction task on \textbf{YT (cmt.)}, we use the provided user-level ideology annotation.
    For user-level prediction on \textbf{YT (user)}, we concatenate all comments by a user.

    \item \texttt{Twitter} \citep[\textbf{TW;}][]{preotiuc-pietro-etal-2017-beyond}: We crawl recent tweets by each user and remove replies and non-English tweets. We assume users' ideologies do not change after their self-report since prior work has shown that people's ideology is less likely to change across the political spectrum \citep{doi:10.1146/annurev.polisci.11.053106.153836}. We sort all tweets from a user chronologically and concatenate them.
\end{itemize}

\begin{table}
\centering
\resizebox{1.0\linewidth}{!}
{%
\begin{tabular}{lccc}
\toprule
\multicolumn{1}{c}{\multirow{2}{*}{Models}}               & \multicolumn{3}{c}{SemEval (Seen)}               \\
\cmidrule(lr){2-4}
 \multicolumn{1}{c}{}             & $\text{F}_{favor}$         & $\text{F}_{against}$       & $\text{F}_{avg}$       \\
\midrule
WKNN \citep{Ghadir_Azmi}          & \textbf{84.49} & 68.36          & \textbf{76.45}      \\
PNEM \citep{Siddiqua_Chy}          & 66.56          & \textbf{77.66}          & 72.11                    \\
MITRE \citep{MITRE}      & 59.32          & 76.33         & 67.82               \\
pkudblab \citep{pkudblab}      & 61.98          & 72.67          & 67.33               \\
SVM-ngrams \citep{semeval}    & 62.98          & 74.98          & 68.98                \\
Majority class \citep{semeval} & 52.01          & 78.44          & 65.22                \\
\midrule
BERT           & 62.89          & 70.75          & 66.82          \\
RoBERTa       & 67.33          & 75.52          & 71.43          \\
\model         & 67.36          & 75.29          & 71.33           \\
\bottomrule
\end{tabular}
}
\caption{Comparison with the state-of-the-art models on SemEval. Prior work (top panel) trains five models (one per target label). On the contrary, in this work, we target a more generalizable approach, i.e., one unified classifier for all labels. Due to different setups, \model and baselines like RoBERTa perform worse.}
\label{table:sota_semeval}
\vspace{-10pt}
\end{table}

\begin{table*}
\centering

\resizebox{1.0\linewidth}{!}{%
\begin{tabular}{lcccccccccccccc}
\toprule
\multicolumn{1}{c}{\multirow{3}{*}{\textbf{}}} & \multicolumn{7}{c}{\textbf{Ideology Prediction}} & \multicolumn{5}{c}{\textbf{Stance Detection}} & \multicolumn{1}{c}{} & \multicolumn{1}{c}{\multirow{3}{*}{\pbox{3em}{All \\ avg}}} \\
\cmidrule(lr){2-8}
\cmidrule(lr){9-14}

\multicolumn{1}{c}{} & \multicolumn{1}{c}{YT} & \multicolumn{1}{c}{\multirow{2}{*}{CongS}} & \multicolumn{1}{c}{\multirow{2}{*}{HP}} & \multicolumn{1}{c}{\multirow{2}{*}{AllS}} & \multicolumn{1}{c}{YT} & \multicolumn{1}{c}{\multirow{2}{*}{TW}} & \multicolumn{1}{c}{Ideo.}  & \multicolumn{1}{c}{SEval} & \multicolumn{1}{c}{SEval} & \multicolumn{1}{c}{Basil} & \multicolumn{1}{c}{\multirow{2}{*}{VAST}} & \multicolumn{1}{c}{Basil} & \multicolumn{1}{c}{Stan.}  & \multicolumn{1}{c}{} \\
\multicolumn{1}{c}{} & \multicolumn{1}{c}{(cmt.)} & \multicolumn{1}{c}{} & \multicolumn{1}{c}{} & \multicolumn{1}{c}{} & \multicolumn{1}{c}{(user)} & \multicolumn{1}{c}{} & \multicolumn{1}{c}{avg} & \multicolumn{1}{c}{(seen)} & \multicolumn{1}{c}{(unseen)} & \multicolumn{1}{c}{(sent.)} & \multicolumn{1}{c}{} & \multicolumn{1}{c}{(art.)} & \multicolumn{1}{c}{avg} & \multicolumn{1}{c}{} \\
\midrule
\multicolumn{15}{l}{\textbf{Baselines}} \\
SVM & 65.34$_{\pm0.00}$ & {\ul 71.31}$_{\pm0.00}$ & 61.25$_{\pm0.00}$ & 52.51$_{\pm0.00}$ & 66.49$_{\pm0.00}$ & 42.85$_{\pm0.00}$ & 59.96 & 51.18$_{\pm0.00}$ & 32.89$_{\pm0.00}$ & 51.08$_{\pm0.00}$ & 39.54$_{\pm0.00}$ & 30.77$_{\pm0.00}$ & 41.09 & 51.38 \\
BERT & 64.64$_{\pm1.92}$ & 65.88$_{\pm1.13}$ & 48.42$_{\pm1.44}$ & 60.88$_{\pm0.83}$ & 65.24$_{\pm1.53}$ & 44.20$_{\pm2.03}$ & 58.21 & 65.07$_{\pm1.02}$ & 40.39$_{\pm0.53}$ & 62.81$_{\pm3.95}$ & 70.53$_{\pm0.43}$ & 45.61$_{\pm3.92}$ & 56.88 & 57.61 \\
RoBERTa & 66.72$_{\pm0.85}$ & 67.25$_{\pm0.48}$ & 60.43$_{\pm3.13}$ & 74.75$_{\pm1.26}$ & 67.98$_{\pm4.03}$ & 48.90$_{\pm1.53}$ & 64.34 & \textbf{70.15}$_{\pm0.87}$ & 63.08$_{\pm0.77}$ & 68.16$_{\pm2.55}$ & 76.25$_{\pm0.11}$ & 41.36$_{\pm7.35}$ & 63.80 & 64.09 \\

\midrule
\multicolumn{15}{l}{\textbf{\newcite{Baly_Martino}}} \\
with Original Data & 65.42$_{\pm0.56}$ & 66.74$_{\pm1.33}$ & 58.37$_{\pm1.63}$ & 72.89$_{\pm0.50}$ & 70.47$_{\pm1.77}$ & 44.95$_{\pm1.02}$ & 63.14 & 68.66$_{\pm0.50}$ & 56.29$_{\pm2.07}$ & 61.30$_{\pm2.41}$ & 75.57$_{\pm0.67}$ & 37.98$_{\pm3.43}$ & 59.96 & 61.69 \\
with \datad & 68.57$_{\pm1.02}$ & 70.39$_{\pm0.38}$ & \textbf{71.24}$_{\pm2.06}$ & {\ul 76.47}$_{\pm3.35}$ & 74.74$_{\pm1.63}$ & 47.38$_{\pm1.31}$ & {\ul 68.13} & 65.84$_{\pm0.74}$ & 49.54$_{\pm2.03}$ & 60.60$_{\pm6.55}$ & 75.03$_{\pm1.58}$ & 41.84$_{\pm9.30}$ & 58.57 & 63.79 \\

\midrule
\multicolumn{15}{l}{\textbf{Our models with triplet loss objective only}} \\

Ideology Obj. & 66.20$_{\pm1.46}$ & \colorbox{cyan!15}{68.18$_{\pm0.54}$} & \colorbox{cyan!15}{64.15$_{\pm6.82}$} & \colorbox{cyan!15}{\textbf{76.52}$_{\pm1.62}$} & \colorbox{cyan!15}{68.15$_{\pm6.89}$} & 42.66$_{\pm10.84}$ & 64.31 & 68.78$_{\pm0.79}$ & 59.61$_{\pm3.97}$ & 64.18$_{\pm4.41}$ & 76.03$_{\pm0.32}$ & \colorbox{cyan!15}{44.94$_{\pm5.61}$} & 62.71 & 63.58 \\
Story Obj. & 66.09$_{\pm1.05}$ & \colorbox{cyan!15}{69.11$_{\pm1.21}$} & 56.70$_{\pm2.64}$ & 74.59$_{\pm1.68}$ & \colorbox{cyan!15}{68.89$_{\pm3.18}$} & 46.53$_{\pm3.29}$ & 63.65 & 69.02$_{\pm0.38}$ & \colorbox{cyan!15}{\textbf{63.54}$_{\pm1.19}$} & 67.21$_{\pm2.51}$ & \colorbox{cyan!15}{76.66$_{\pm1.29}$} & \colorbox{cyan!15}{\textbf{53.16}$_{\pm6.76}$} & \colorbox{cyan!15}{{\ul 65.92}} & \colorbox{cyan!15}{64.68} \\
Ideology Obj. + Story Obj. & \colorbox{cyan!15}{{\ul 68.91}$_{\pm0.44}$} & \colorbox{cyan!15}{69.10$_{\pm0.71}$} & \colorbox{cyan!15}{63.08$_{\pm3.10}$} & \colorbox{cyan!15}{76.23$_{\pm2.96}$} & \colorbox{cyan!15}{{\ul 77.58}$_{\pm2.83}$} & \colorbox{cyan!15}{\textbf{48.98$_{\pm1.42}$}} & \colorbox{cyan!15}{ 67.31} & 69.66$_{\pm0.45}$ & \colorbox{cyan!15}{{\ul 63.17}$_{\pm1.92}$} & 64.37$_{\pm1.58}$ & 76.18$_{\pm1.13}$ & \colorbox{cyan!15}{47.01$_{\pm7.55}$} & \colorbox{cyan!15}{64.08} & \colorbox{cyan!15}{{\ul 65.84}} \\

\noalign{\vskip 0.3ex}
\hdashline[5pt/4pt]
\noalign{\vskip 0.3ex}
\multicolumn{15}{l}{\textbf{Our models with masked language model objective only}} \\
Random & 67.82$_{\pm1.30}$ & 70.32$_{\pm0.94}$ & 60.59$_{\pm2.22}$ & 73.54$_{\pm1.55}$ & 70.77$_{\pm1.43}$ & 44.62$_{\pm2.32}$ & 64.61 & 69.16$_{\pm0.84}$ & 60.39$_{\pm0.85}$ & 69.94$_{\pm1.61}$ & \textbf{77.11}$_{\pm0.53}$ & 39.16$_{\pm3.71}$ & 63.15 & 63.95 \\
Upsamp. Ent. & \colorbox{green!15}{\textbf{69.06}$_{\pm1.00}$} & \colorbox{green!15}{70.32$_{\pm0.39}$} & 60.09$_{\pm0.98}$ & 70.89$_{\pm1.81}$ & \colorbox{green!15}{71.40$_{\pm2.23}$} & \colorbox{green!15}{47.16$_{\pm1.07}$} & \colorbox{green!15}{64.82} & \colorbox{green!15}{69.81$_{\pm0.61}$} & \colorbox{green!15}{63.08$_{\pm1.90}$} & 69.49$_{\pm1.85}$ & 76.76$_{\pm1.01}$ & \colorbox{green!15}{46.46$_{\pm5.56}$} & \colorbox{green!15}{65.12} & \colorbox{green!15}{64.96} \\
Upsamp. Sentiment & 67.41$_{\pm1.12}$ & 70.03$_{\pm0.96}$ & 56.05$_{\pm5.68}$ & 72.35$_{\pm1.09}$ & \colorbox{green!15}{74.93$_{\pm2.70}$} & \colorbox{green!15}{48.15$_{\pm1.30}$} & \colorbox{green!15}{64.82} & \colorbox{green!15}{{\ul 70.09}$_{\pm0.51}$} & \colorbox{green!15}{60.81$_{\pm1.22}$} &  \colorbox{green!15}{{\ul 71.28}$_{\pm2.31}$} & 76.61$_{\pm0.62}$ & \colorbox{green!15}{44.42$_{\pm4.91}$} & \colorbox{green!15}{64.64} & \colorbox{green!15}{64.74} \\
Upsamp. Ent. + Sentiment & \colorbox{green!15}{68.31$_{\pm0.37}$} & \colorbox{green!15}{\textbf{71.42}$_{\pm0.51}$} & 58.02$_{\pm3.34}$ & 71.90$_{\pm0.61}$ & \colorbox{green!15}{71.04$_{\pm3.56}$} & \colorbox{green!15}{47.31$_{\pm2.07}$} & \colorbox{green!15}{64.67} & \colorbox{green!15}{69.25$_{\pm0.71}$} & \colorbox{green!15}{62.84$_{\pm3.93}$} & 69.23$_{\pm1.08}$ & {\ul 77.10}$_{\pm0.73}$ & \colorbox{green!15}{43.16$_{\pm4.95}$} & \colorbox{green!15}{64.32} & \colorbox{green!15}{64.51} \\

\noalign{\vskip 0.3ex}
\hdashline[5pt/4pt]
\noalign{\vskip 0.3ex}
\model & \colorbox{magenta!15}{67.83$^\ast$$_{\pm0.49}$} & 70.86$_{\pm0.31}$ & \colorbox{magenta!15}{{\ul 70.25}$^\ast$$_{\pm2.10}$} & \colorbox{magenta!15}{74.93$_{\pm0.83}$} & \colorbox{magenta!15}{\textbf{78.73}$^\ast$$_{\pm1.15}$} & \colorbox{magenta!15}{{\ul 48.92}$_{\pm2.19}$} & \colorbox{magenta!15}{\textbf{68.59}} & 69.41$_{\pm0.36}$ & 61.26$_{\pm1.23}$ & \colorbox{magenta!15}{\textbf{73.41$^\ast$}$_{\pm0.97}$} & \colorbox{magenta!15}{76.73$^\ast$$_{\pm0.60}$} & \colorbox{magenta!15}{{\ul 51.94}$^\ast$$_{\pm3.42}$} & \colorbox{magenta!15}{\textbf{66.55}} & \colorbox{magenta!15}{\textbf{67.66}} \\

\bottomrule

\end{tabular}
}
\caption{
Macro F1 scores on 11 evaluation tasks (average of 5 runs) with standard deviations. 
Tasks are sorted by text length, short to long, within each group. 
``All avg'' is the average of all 11 tasks. 
\textbf{Best} results are in bold and \underline{second best} are underlined. 
Our models with triplet-loss objectives that outperform RoBERTa are in \colorbox{cyan!15}{blue}. 
Our models with specialized sampling methods that outperform vanilla MLM (Random) are in \colorbox{green!15}{green}. 
\model uses Ideology + Story Obj. and Upsamp. Ent. + Sentiment. 
Results where \model outperforms all baselines are highlighted in \colorbox{magenta!15}{red}. 
\model has the smallest standard deviations on 5 tasks, showing its stable performance.
}
\vspace{-10pt}
\label{table:results_new}
\end{table*}

\subsection{Stance Detection}
\label{app:stance_detection}
\begin{itemize}[leftmargin=1em,noitemsep,topsep=0pt,parsep=0pt,partopsep=0pt]
    \item \texttt{BASIL}\footnote{\url{https://github.com/marshallwhiteorg/emnlp19-media-bias}.} \citep{basil}: We convert the original dataset such that the new tasks are to predict the stance towards a target at two granularities: article (\textbf{art.}) and sentence (\textbf{sent.}) levels. The targets in the dataset can be a person (e.g., Donald Trump) or an organization (e.g., Justice Department).

    \item \texttt{VAST}\footnote{\url{https://github.com/emilyallaway/zero-shot-stance}.} \citep{allaway-mckeown-2020-zero} predicts the stance of a comment towards a target. The targets in the dataset are noun phrases covering a broad range of topics (e.g., immigration, home schoolers).
    We notice the original dataset contains contradictory samples, where the same comment-target pair is annotated with opposite stances, and therefore remove duplicate and contradictory samples.
    
    \item \texttt{SemEval}\footnote{\url{https://alt.qcri.org/semeval2016/task6/index.php?id=data-and-tools}.} \citep[\textbf{SEval;}][]{semeval} predicts a tweet's stance towards a target.
    The dataset contains six targets: Atheism, Climate Change, Feminist, Hillary Clinton, Abortion, and Donald Trump.
    Notably, the last target is not seen during training, and only appears in testing.
\end{itemize}

\section{Task Property}
\label{appendix:property}
This section introduces detailed definitions of four properties, based on which we divide tasks into two categories reported in Figure~\ref{fig:grouping}.
\begin{itemize}[leftmargin=1em,noitemsep,topsep=0pt,parsep=0pt,partopsep=0pt]
    \item Formality: Speech and news genres are considered as formal, and others are informal.
    \item Training set size: Datasets with more than 2,000 training samples are categorized as large, and small otherwise.
    \item Document length: Datasets with average document length larger than 500 are treated as ``long'', and others are short.
    \item Aggregation level: If a dataset is a collection of single articles/posts/tweets, then it is in the category of ``Single''. If posts are concatenated and aggregated at user level, then it is marked as  ``User''. Specifically, only \texttt{YouTube User} and \texttt{Twitter} in Table~\ref{table:benchmark} are in the ``User'' category.  
\end{itemize}

\section{Comparison with Previous State-of-the-art Models}
\label{app:SOTA_comparison}
Here we compare \model with previous state-of-the-art models on three selected datasets: \texttt{Hyperpartisan} (\cref{section:results}), \texttt{VAST} (\cref{app:vast}), and \texttt{SemEval} (\cref{app:seval}). \cref{app:inapplicable} discusses why direct comparisons are not applicable on the other 5 datasets.

\subsection{VAST}
\label{app:vast}

\model outperforms all previous state-of-the-art models in the literature as well as the strong RoBERTa baseline, as can be seen in Table~\ref{table:sota_vast}. Following \citet{allaway-mckeown-2020-zero} and \citet{Jayaram_Allaway}, F$_{avg}$ is defined as the macro-averaged F1 over all three classes (favor, against, and neutral). The results are reported on the original VAST dataset which contains contradictory samples, where the same comment-target pairs are annotated with opposite stances so they are counted in both categories.

\subsection{SemEval}
\label{app:seval}
Table~\ref{table:sota_semeval} show the results of state-of-the-art models and \model on SemEval.
Following \citet{semeval}, F$_{avg}$ is defined as the macro-averaged F1 over favor and against classes. State-of-the-art models train separate classifiers, one for each target, thus yield better results than \model. 
Similar observation is made by \citet{semeval}, where one single SVM that is trained on all five targets performs worse than five one-versus-rest SVM classifiers.

\subsection{Reasons for Inapplicable Comparisons}
\label{app:inapplicable}
We are unable to directly compare with existing models on datasets other than \texttt{Hyperpartisan}, \texttt{VAST}, and \texttt{SemEval} for the following  reasons:
\begin{itemize}[leftmargin=1em,noitemsep,topsep=0pt,parsep=0pt,partopsep=0pt]
    \item The original dataset either is used for different tasks that are not ideology prediction or stance detection: \texttt{Congress Speech} \citep{congress-speech}, \texttt{BASIL} \citep{basil}, and \texttt{YouTube} \citep{wu2021crosspartisan}.
    \item The dataset is newly collected (\texttt{AllSides}) or contains newly collected samples 
    \citep[\texttt{Twitter};][]{preotiuc-pietro-etal-2017-beyond}.
\end{itemize}

\section{Visualize Attention Weights}
\label{section:appendix-attention-viz}

\vspace{-15pt}

\begin{figure*}[!htb]
    \centering
    \includegraphics[width=0.95\textwidth]{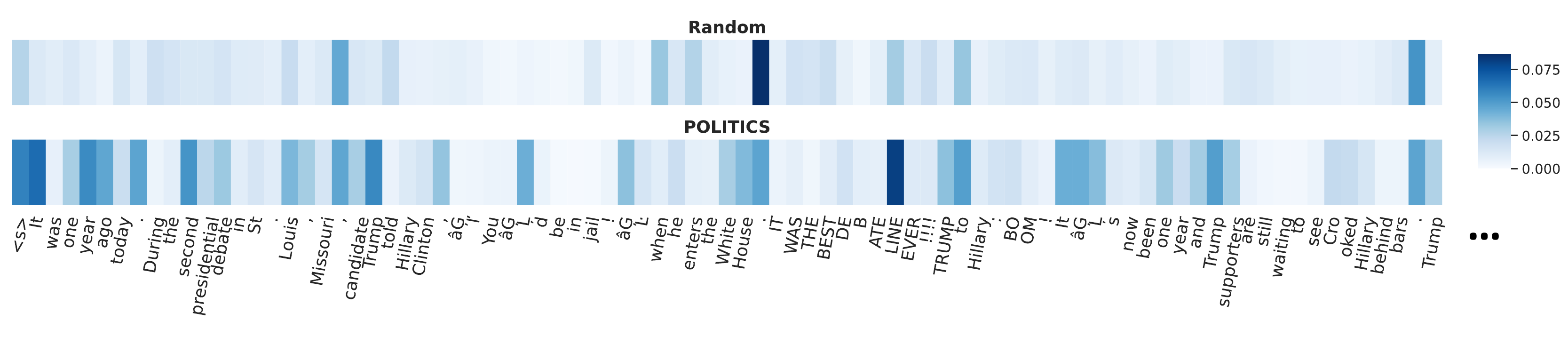}
    \caption{Example 1 for attention visualization. Last layer attention weights between \texttt{[CLS]} token and other tokens in the input. We illustrate the first 85 tokens of the article.}
    \label{fig:append-atten-debate}
\end{figure*}

\vspace{-15pt}

\vspace{-15pt}

\begin{figure*}[!htb]
    \centering
    \includegraphics[width=0.95\textwidth]{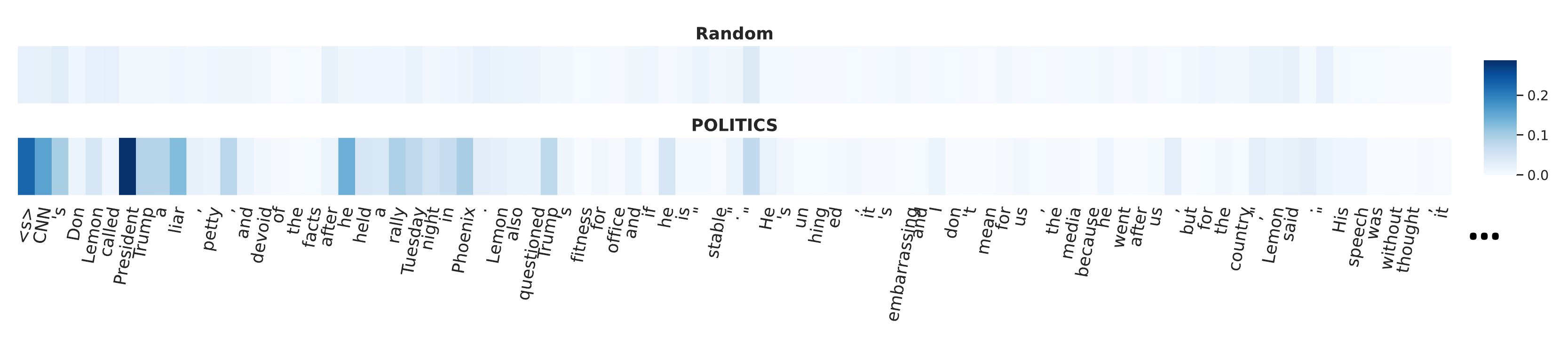}
    \vspace{-15pt}
    \caption{Example 2 for attention visualization. Last layer attention weights between \texttt{[CLS]} token and other tokens in the input. We illustrate the first 85 tokens of the article.}
    \label{fig:append-atten-cnn}
\end{figure*}

\vspace{-15pt}

\begin{figure*}[!htb]
    \centering
    \includegraphics[width=0.95\textwidth]{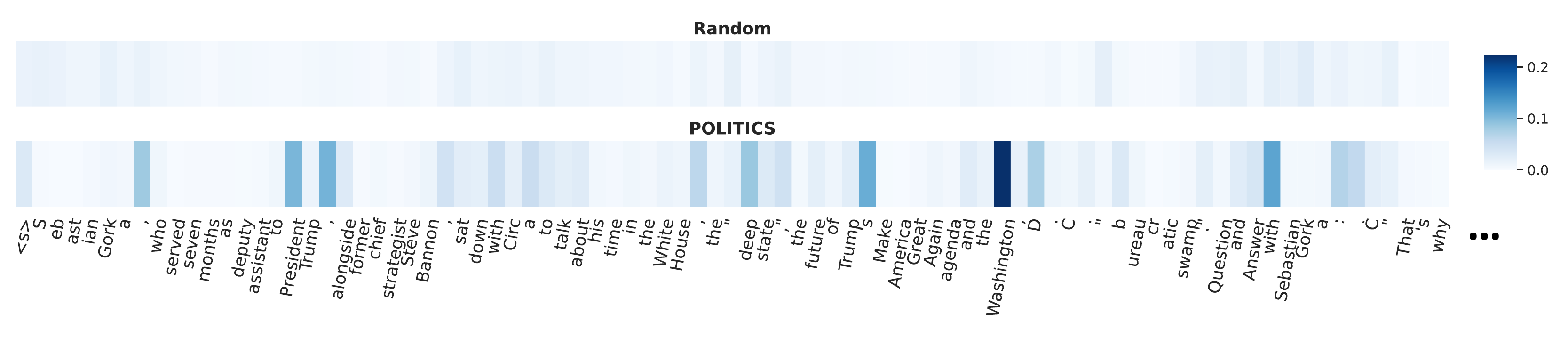}
    \vspace{-15pt}
    \caption{Example 3 for attention visualization. Last layer attention weights between \texttt{[CLS]} token and other tokens in the input. We illustrate the first 85 tokens of the article.}
    \label{fig:append-atten-seb}
\end{figure*}

\vspace{-15pt}

\begin{figure*}[!htb]
    \centering
    \includegraphics[width=0.95\textwidth]{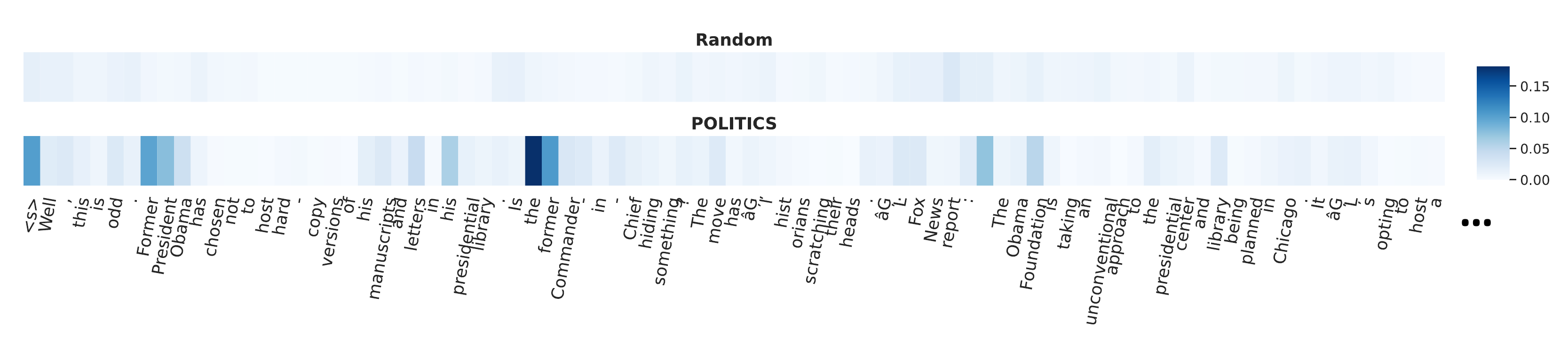}
    \vspace{-15pt}
    \caption{Example 4 for attention visualization. Last layer attention weights between \texttt{[CLS]} token and other tokens in the input. We illustrate the first 85 tokens of the article.}
    \label{fig:append-atten-obama}
\end{figure*}

\end{appendices}

\appendix

\end{document}